\documentclass{article}

 \usepackage[preprint]{neurips_2025}

 \usepackage[utf8]{inputenc} 
\usepackage[T1]{fontenc}    
\usepackage{hyperref}       
\usepackage{url}            
\usepackage{booktabs}       
\usepackage{amsfonts}       
\usepackage{nicefrac}       
\usepackage{microtype}      
\usepackage{xcolor}         

\usepackage{floatflt}
\usepackage{float}
\usepackage{wrapfig}
\usepackage{subcaption}

\usepackage{times}
\usepackage{latexsym}

\usepackage[T1]{fontenc}

\usepackage[utf8]{inputenc}

\usepackage{microtype}

\usepackage{inconsolata}

\usepackage{graphicx}
\usepackage{multirow}
\usepackage[table,xcdraw]{xcolor}

\usepackage{xcolor} 
\usepackage{soul}   
\usepackage{tikz}   
\usepackage{pgfplots}
\pgfplotsset{compat=1.18}

\usepackage{inconsolata}
\usepackage{listing}
\usepackage{minted}

\usepackage{multirow}
\usepackage[table,xcdraw]{xcolor}

\title{The Oracle Has Spoken: A Multi-Aspect\\ Evaluation of Dialogue in Pythia}

\author{%
  Zixun Chen \\
  Columbia University\\
  New York, NY \\
  \texttt{zc2738@columbia.edu} \\
  \And
  Petr Babkin\thanks{Corresponding Author.} \\
  J.P. Morgan AI Research \\ 
  Palo Alto, CA \\
  \texttt{petr.babkin@jpmorgan.com} \\
  \And
  Akshat Gupta \\
  University of California, Berkeley 
  \\ Berkeley, CA \\
  \texttt{akshat.gupta@berkeley.edu} \\ \\
  \And
  Gopala Anumanchipalli \\
  University of California, Berkeley \\ 
  Berkeley, CA \\
  \texttt{gopala@berkeley.edu} \\
  \And
  Xiaomo Liu \\
  J.P. Morgan AI Research \\ 
  New York, NY \\
  \texttt{xiaomo.liu@jpmchase.com}
}

\begin{document}
\maketitle
\begin{abstract}
Dialogue is one of the landmark abilities of large language models (LLMs).
Despite its ubiquity, few studies actually distinguish specific ingredients underpinning  dialogue behavior emerging during post-training.
We employ a comprehensive suite of model-based metrics, each targeting a distinct fine-grained aspect of dialogue, motivated by linguistic theory. 
We evaluate how the performance of pre-trained Pythia models changes with respect to each of those dimensions, depending on model size and as a result of supervised fine-tuning on conversational datasets.
We observe only a mild impact of raw model size on most metrics, whereas fine-tuning quickly saturates the scores for all but the smallest models tested.
Somewhat contrary to our expectations, many metrics show very similar trends, especially if they are all rooted in the same evaluator model, which raises the question of their reliability in measuring a specific dimension.
To that end, we conduct additional analyses of score distributions, metric correlations, and term frequencies in generated responses to help explain our observations.
\end{abstract}

\section{Introduction}
Large Language Models (LLMs) took the world by storm, in part, thanks to their ability to engage in naturalistic dialogue. 
While the acquisition of general language fluency and world knowledge by LLMs is commonly attributed to the phase of large scale self-supervised pre-training \citep{brown2020languagemodelsfewshotlearners}, pinpointing the origins of specific abilities, such as dialogue, remains a challenge.
Although it is tempting to take dialogue abilities of LLMs for granted, as naturally emerging from pre-training, 
there is a well documented behavior of raw base models responding to a prompt by reproducing its likely continuation in the pre-training corpus rather than conversing in a way a human person would\footnote{\url{https://openai.com/index/instruction-following/}}.
A comparatively small-scale phase of post-training, which includes supervised fine-tuning on human-annotated prompt/response pairs, proved essential to ensuring alignment with user intent and preferences --- including making them more conversational \citep{ouyang2022training}. 
Fine-tuning is a well known technique for improving downstream task performance.
Due to its significantly smaller scale, fine-tuning merely ``activates'' skills and knowledge already learned during pre-training, and aligns them with user preferences; and its effectiveness is upper bounded by the base model's size
\citep{zhou2023lima}. 
Similarly, LLM performance on downstream tasks is largely determined by the amount of pre-training of the base model \citep{gudibande2023false}. 
Furthermore, fine-tuning of smaller models, using imitation data from strong LLMs, teaches them to mimic the style of larger models but is generally unable to compensate for limited pre-training knowledge. 


Whether or not an LLM's ability to engage in a conversation merely boils down to stylistic adaptation or constitutes a skill in and of itself, it is clear from linguistic theory that dialogue is predicated on the ability to recognize intent, infer discourse relations between utterances, and to keep track of the evolving state of the conversation \citep{80dc171d49614287b324a0d2b0d201fc}.

In this work, we investigate the effects of conversational finetuning on an open source LLM Pythia \citep{biderman2023pythia} of five different sizes through the lens of these fine-grained linguistic dimensions.
We contribute an empirical analysis of changes in model's text generation behavior, as a result of finetuning on 3 chat datasets: Databricks Dolly\footnote{\url{https://huggingface.co/datasets/databricks/databricks-dolly-15k}}, Open Assistant\footnote{\url{https://huggingface.co/datasets/OpenAssistant}}, and ShareGPT\footnote{\url{https://huggingface.co/datasets/anon8231489123/ShareGPT_Vicuna_unfiltered}}, with respect to dialogue-specific metrics: UniEval \citep{zhong2022unified}, Themis \citep{hu-etal-2024-themis}, and a targeted GPT-4-based assessment.

\section{Related Work}
\label{sec:related-work}

Several works previously evaluated LLMs fine-tuned on conversational data.
We summarize them below, while focusing on differences from our work.

Alghisi et al. \citeyearpar{alghisi2024finetuneragevaluatingdifferent} evaluated different LLM adaptation techniques: in-context learning, fine-tuning and RAG across open domain, knowledge-grounded and task-based dialogue. They find fine-tuning to provide superior lift compared to ICL while retrieval augmentation moderately improves both ICL and SFT. 
Notably, they used chat/instruct variants of Llama \citep{touvron2023llama2openfoundation} and Mistral \citep{jiang2023mistral7b} without controlling for model size. Whereas, we start with base models of different sizes and directly fine-tune them on each dataset without relying on chat/instruct variants. 
Additionally the use of perplexity for evaluation is limiting, as it gives an overall measure of gold text's uncertainty under the model, whereas we look at fine-grained linguistic dimensions with greater interpretability.

Mousavi et al. \citeyearpar{mousavi-etal-2022-evaluation} also conducted dialogue fine-tuning experiments with T5 \citep{raffel2023exploringlimitstransferlearning} and GPT-2 \citep{Radford2019LanguageMA} in Italian. 
Their work emphasized human evaluation of dimensions such as appropriateness, contextualization, and grammar correctness, which are more informative than perplexity but are quite ad hoc (e.g., ``Genericness'') and are not based on any concrete linguistic phenomena. 
Despite our evaluation using automatic metrics, they are all based on models specifically trained to reproduce human judgements on text annotation tasks.

DialogBench \citep{ou2024dialogbenchevaluatingllmshumanlike} is a synthetic dataset of dialogue-related tasks encompassing intent recognition, knowledge grounded generation and coherent infilling, rendered as multiple-choice questions. 
By contrast, we use organic conversations for which we simultaneously compute multiple metrics each targeting a different aspect of dialogue.
Authors evaluated a large number of pre-trained and chat/instruction-tuned LLMs from different families and sizes and noted base LLMs do well on correctness-related tasks but struggle with coherence and safety. Our evaluation additionally includes analyses of metric reliability.




\section{Methodology}
\label{sec:method}



We use the Pythia family of models \citep{biderman2023pythia} as the testbed for our fine-tuning experiments. 
These decoder-only LLMs are pre-trained on 1 trillion tokens of the Pile corpus and come in 10 model sizes, ranging from 14 million to 12 billion parameters --- all trained on data presented in the same order to facilitate reproducibility research. To reasonably cover the configuration space, we select the following checkpoints: 140m, 410m, 1.3b, 2.8b, 6.9b (all dedup).

To ensure diversity of the the fine-tuning datasets, we use a combination of single and multi-turn corpora authored by human annotators, as well as synthesized by ChatGPT: Dolly, Open Assistant, ShareGPT. We sample 10k conversations from each.

\subsection{Evaluation Metrics}
Evaluation of dialogue is inherently challenging due to a large space of valid responses to any given utterance. Rather than capturing general quality of generated responses (which is expected to improve through fine-tuning on any corpus), our goal is to disentangle specific linguistic dimensions that contribute to high-quality dialogue. 
While human evaluation is considered a gold standard for assessing LLMs, it comes with its own set of challenges, including bias, subjectivity, low reproducibility, and high cost.
After an extensive survey of automatic metrics, we adopt the following 3 model-based metrics that most closely reflect qualities characteristic of dialogue.

\paragraph{UniEval} \citep{zhong2022unified}
is based on the T5 model, finetuned in the to predict entailment between a text and a set of interpretable attributes, which for dialogue include: Naturalness, Coherence, Engagingness, Groundedness, and Understandability. Depending on a metric, it can be computed either on the evaluated model's output alone (e.g., Naturalness and Understandability), model's output in conjunction with 
either the prompt (e.g., Coherence), or an optional context (e.g., Groundedness), and finally all 3 (e.g., Engagingness). 


\paragraph{Themis} \citep{hu-etal-2024-themis} is a more recent model-based evaluation created by finetuning Llama3-8B \citep{grattafiori2024llama3herdmodels} on 58 datasets, encompassing 9 generative language tasks, with additional preference alignment and multi-perspective consistency validation. 
Themis evaluates the following 4 aspects for dialogue: Context Maintenance, Interestingness,
Knowledge Use, Naturalness. In addition to scores on the 5-point scale, Themis provides reviews to back up its ratings.

\paragraph{GPT-4-as-a-judge.} Evaluation through prompting a strong LLM, such as GPT-4 \citep{openai2024gpt4technicalreport} has a emerged as a new paradigm in NLP. We use this as an opportunity to fine-tune evaluated dimensions to probe nuanced pragmatic aspects of dialogue not usually captured in generic evaluations.
Specifically, we instructed GPT-4 to assess how well our LLMs are able to follow dialogue turn taking conventions, recognize user's intention, generate coherent responses, and keep track of discourse referents. The complete prompt used in the FastChat format\footnote{\url{https://github.com/lm-sys/FastChat}} is included in the appendix.

We manually examined 5-10 lowest and highest scoring examples from the datasets for each metric, concluding they reasonably captured our intended evaluation dimensions (see Sections \ref{sec:qualitative-unieval}--\ref{sec:qualitative-gpt4}). 
We conclude by and large UniEval, Themis, and GPT-4-based metrics are reliable indicators of generation quality with respect to the evaluated dimensions.
Nonetheless, it should be noted, a high score in any of these metrics does not guarantee the generations to be completely sensible or free of obvious hallucinations.
In addition, we have come across a small number of clearly incoherent examples that still obtained high scores in UniEval. 



\paragraph{Open LLM Leaderboard.} While probing the models for dialogue specific competencies, we also track the average of the common NLP benchmarks, such as MMLU, TruthfulQA, and Winogrande used in the Open LLM Leaderboard\footnote{\url{https://huggingface.co/spaces/HuggingFaceH4/open_llm_leaderboard}} for ease of comparison --- to serve as the base level of model's task-specific performance (Appendix \ref{sec:openllm}).

\section{Results and Discussion}
\label{sec:evaluation}
We have run finetuning experiments using the Lit-GPT library\footnote{\url{https://github.com/Lightning-AI/litgpt}} on two NVIDIA RTX A6000's for model sizes from 160 million to 2.8 billion parameters, and four A6000's for the 6.9 billion parameter model. PyTorch Fully-Sharded Data Parallelism \citep{zhao2023pytorch} is enabled for the finetuning of 2.8b and 6.9b parameter models, allowing us to partition weights between GPUs.
Depending on the base model size, we used a learning rate of $3 \times 10^{-5}$ for the 160m model, $1 \times 10^{-5}$ for the 410m model, $1 \times 10^{-6}$ for all other models with the AdamW \citep{loshchilov2019decoupled} optimizer, which we empirically verified to result in best validation set performance across all datasets. We used a batch size of 128 and fine-tuned for 10 epochs for each experiment, keeping the checkpoint with lowest validation loss as well as the base and final checkpoints for verification.
We initially considered parameter-efficient finetuning using the adapter approach \citep{hu2023llmadapters}, however it did not provide a tangible lift, therefore in the experiments reported we used full finetuning.
All reported metrics have been computed on the held out test set of each respective dataset.

\begin{table}[]
\caption{Dialogue metrics for each model size before and after fine-tuning.}
\label{tab:results}
\begin{tabular}{ll|llll|llll|llll}
\hline
 &
   &
  \multicolumn{4}{c}{\textbf{UniEval}} &
  \multicolumn{4}{c}{\textbf{Themis}} &
  \multicolumn{4}{c}{\textbf{GPT-4}} \\ \hline
 &
   &
  \rotatebox{90}{Naturalness} &
  \rotatebox{90}{Coherence} &
  \rotatebox{90}{Groundedness} &
  \rotatebox{90}{Understandability} &
  \rotatebox{90}{Context Maintenance} &
  \rotatebox{90}{Interestingness} &
  \rotatebox{90}{Knowledge Use} &
  \rotatebox{90}{Naturalness} &
  \rotatebox{90}{Turn Taking} &
  \rotatebox{90}{Intent Recognition} &
  \rotatebox{90}{Rhetoric Structure} &
  \rotatebox{90}{Reference Resolution} \\ \hline
\multicolumn{1}{r}{} &
  160m &
  \multicolumn{1}{r}{\cellcolor[HTML]{FFFFFF}0.34} &
  \multicolumn{1}{r}{\cellcolor[HTML]{FFFFFF}0.58} &
  \multicolumn{1}{r}{\cellcolor[HTML]{B3E1CA}0.73} &
  \multicolumn{1}{r}{\cellcolor[HTML]{FFFFFF}0.33} &
  \multicolumn{1}{r}{\cellcolor[HTML]{FFFFFF}1.00} &
  \multicolumn{1}{r}{\cellcolor[HTML]{FFFFFF}1.00} &
  \multicolumn{1}{r}{\cellcolor[HTML]{FFFFFF}1.00} &
  \multicolumn{1}{r}{\cellcolor[HTML]{FFFFFF}1.00} &
  \multicolumn{1}{r}{\cellcolor[HTML]{FFFFFF}1.02} &
  \multicolumn{1}{r}{\cellcolor[HTML]{FFFFFF}1.00} &
  \multicolumn{1}{r}{\cellcolor[HTML]{FFFFFF}1.01} &
  \multicolumn{1}{r}{\cellcolor[HTML]{FFFFFF}1.00} \\
\multicolumn{1}{r}{} &
  410m &
  \multicolumn{1}{r}{\cellcolor[HTML]{E8F6EF}0.39} &
  \multicolumn{1}{r}{\cellcolor[HTML]{88CFAC}0.85} &
  \multicolumn{1}{r}{\cellcolor[HTML]{63C092}0.85} &
  \multicolumn{1}{r}{\cellcolor[HTML]{E5F5ED}0.40} &
  \multicolumn{1}{r}{\cellcolor[HTML]{FEFFFE}1.03} &
  \multicolumn{1}{r}{\cellcolor[HTML]{F8FDFA}1.14} &
  \multicolumn{1}{r}{\cellcolor[HTML]{FBFEFC}1.09} &
  \multicolumn{1}{r}{\cellcolor[HTML]{FDFFFE}1.04} &
  \multicolumn{1}{r}{\cellcolor[HTML]{F9FDFB}1.24} &
  \multicolumn{1}{r}{\cellcolor[HTML]{FDFFFE}1.09} &
  \multicolumn{1}{r}{\cellcolor[HTML]{FCFEFD}1.13} &
  \multicolumn{1}{r}{\cellcolor[HTML]{FDFEFE}1.10} \\
\multicolumn{1}{r}{} &
  1.4b &
  \multicolumn{1}{r}{\cellcolor[HTML]{FAFDFC}0.35} &
  \multicolumn{1}{r}{\cellcolor[HTML]{A3DABF}0.79} &
  \multicolumn{1}{r}{\cellcolor[HTML]{8ED2B1}0.78} &
  \multicolumn{1}{r}{\cellcolor[HTML]{F9FDFB}0.35} &
  \multicolumn{1}{r}{\cellcolor[HTML]{FDFEFE}1.06} &
  \multicolumn{1}{r}{\cellcolor[HTML]{F5FBF8}1.21} &
  \multicolumn{1}{r}{\cellcolor[HTML]{FAFDFC}1.11} &
  \multicolumn{1}{r}{\cellcolor[HTML]{FCFEFD}1.06} &
  \multicolumn{1}{r}{\cellcolor[HTML]{FDFEFE}1.11} &
  \multicolumn{1}{r}{\cellcolor[HTML]{FEFFFE}1.06} &
  \multicolumn{1}{r}{\cellcolor[HTML]{FEFFFF}1.06} &
  \multicolumn{1}{r}{\cellcolor[HTML]{FFFFFF}1.03} \\
\multicolumn{1}{r}{} &
  2.8b &
  \multicolumn{1}{r}{\cellcolor[HTML]{C9E9D9}0.46} &
  \multicolumn{1}{r}{\cellcolor[HTML]{81CCA8}0.87} &
  \multicolumn{1}{r}{\cellcolor[HTML]{61BF91}0.85} &
  \multicolumn{1}{r}{\cellcolor[HTML]{C8E9D9}0.47} &
  \multicolumn{1}{r}{\cellcolor[HTML]{ECF8F2}1.39} &
  \multicolumn{1}{r}{\cellcolor[HTML]{DDF1E7}1.69} &
  \multicolumn{1}{r}{\cellcolor[HTML]{E3F4EC}1.55} &
  \multicolumn{1}{r}{\cellcolor[HTML]{F0F9F5}1.31} &
  \multicolumn{1}{r}{\cellcolor[HTML]{EBF7F2}1.79} &
  \multicolumn{1}{r}{\cellcolor[HTML]{EFF9F4}1.65} &
  \multicolumn{1}{r}{\cellcolor[HTML]{F2FAF6}1.52} &
  \multicolumn{1}{r}{\cellcolor[HTML]{F6FCF9}1.39} \\
\multicolumn{1}{r}{\multirow{-5}{*}{\rotatebox{90}{Base}}} &
  6.9b &
  \multicolumn{1}{r}{\cellcolor[HTML]{DCF1E7}0.42} &
  \multicolumn{1}{r}{\cellcolor[HTML]{79C9A2}0.89} &
  \multicolumn{1}{r}{\cellcolor[HTML]{57BB8A}0.87} &
  \multicolumn{1}{r}{\cellcolor[HTML]{D8F0E4}0.43} &
  \multicolumn{1}{r}{\cellcolor[HTML]{F0F9F5}1.30} &
  \multicolumn{1}{r}{\cellcolor[HTML]{DEF2E8}1.67} &
  \multicolumn{1}{r}{\cellcolor[HTML]{E5F5ED}1.52} &
  \multicolumn{1}{r}{\cellcolor[HTML]{F1FAF6}1.28} &
  \multicolumn{1}{r}{\cellcolor[HTML]{EBF7F1}1.82} &
  \multicolumn{1}{r}{\cellcolor[HTML]{ECF8F2}1.75} &
  \multicolumn{1}{r}{\cellcolor[HTML]{E7F6EE}1.96} &
  \multicolumn{1}{r}{\cellcolor[HTML]{F1FAF5}1.56} \\ \hline
\multicolumn{1}{r}{} &
  160m &
  \multicolumn{1}{r}{\cellcolor[HTML]{D9F0E5}0.42} &
  \multicolumn{1}{r}{\cellcolor[HTML]{81CCA7}0.87} &
  \multicolumn{1}{r}{\cellcolor[HTML]{EAF7F0}0.64} &
  \multicolumn{1}{r}{\cellcolor[HTML]{D5EEE2}0.44} &
  \multicolumn{1}{r}{\cellcolor[HTML]{E9F6F0}1.45} &
  \multicolumn{1}{r}{\cellcolor[HTML]{E2F4EB}1.58} &
  \multicolumn{1}{r}{\cellcolor[HTML]{EBF7F1}1.41} &
  \multicolumn{1}{r}{\cellcolor[HTML]{EAF7F1}1.42} &
  \multicolumn{1}{r}{\cellcolor[HTML]{DFF2E9}2.29} &
  \multicolumn{1}{r}{\cellcolor[HTML]{EFF9F4}1.65} &
  \multicolumn{1}{r}{\cellcolor[HTML]{EDF8F3}1.72} &
  \multicolumn{1}{r}{\cellcolor[HTML]{F6FCF9}1.37} \\
\multicolumn{1}{r}{} &
  410m &
  \multicolumn{1}{r}{\cellcolor[HTML]{6EC59A}0.67} &
  \multicolumn{1}{r}{\cellcolor[HTML]{5DBE8F}0.95} &
  \multicolumn{1}{r}{\cellcolor[HTML]{FFFFFF}0.60} &
  \multicolumn{1}{r}{\cellcolor[HTML]{70C59B}0.69} &
  \multicolumn{1}{r}{\cellcolor[HTML]{98D6B8}3.05} &
  \multicolumn{1}{r}{\cellcolor[HTML]{A0D9BD}2.88} &
  \multicolumn{1}{r}{\cellcolor[HTML]{A2DABE}2.85} &
  \multicolumn{1}{r}{\cellcolor[HTML]{9FD9BC}2.91} &
  \multicolumn{1}{r}{\cellcolor[HTML]{97D5B7}5.08} &
  \multicolumn{1}{r}{\cellcolor[HTML]{A7DCC2}4.45} &
  \multicolumn{1}{r}{\cellcolor[HTML]{B7E2CD}3.82} &
  \multicolumn{1}{r}{\cellcolor[HTML]{B7E2CD}3.85} \\
\multicolumn{1}{r}{} &
  1.4b &
  \multicolumn{1}{r}{\cellcolor[HTML]{5EBE8F}0.70} &
  \multicolumn{1}{r}{\cellcolor[HTML]{59BC8B}0.96} &
  \multicolumn{1}{r}{\cellcolor[HTML]{EFF9F4}0.63} &
  \multicolumn{1}{r}{\cellcolor[HTML]{5FBF90}0.74} &
  \multicolumn{1}{r}{\cellcolor[HTML]{6CC499}3.93} &
  \multicolumn{1}{r}{\cellcolor[HTML]{86CEAB}3.41} &
  \multicolumn{1}{r}{\cellcolor[HTML]{73C79E}3.78} &
  \multicolumn{1}{r}{\cellcolor[HTML]{72C69D}3.79} &
  \multicolumn{1}{r}{\cellcolor[HTML]{76C8A0}6.36} &
  \multicolumn{1}{r}{\cellcolor[HTML]{76C8A0}6.37} &
  \multicolumn{1}{r}{\cellcolor[HTML]{8BD0AE}5.56} &
  \multicolumn{1}{r}{\cellcolor[HTML]{87CFAB}5.72} \\
\multicolumn{1}{r}{} &
  2.8b &
  \multicolumn{1}{r}{\cellcolor[HTML]{5ABD8C}0.71} &
  \multicolumn{1}{r}{\cellcolor[HTML]{5ABC8C}0.96} &
  \multicolumn{1}{r}{\cellcolor[HTML]{F9FDFB}0.62} &
  \multicolumn{1}{r}{\cellcolor[HTML]{5BBD8D}0.75} &
  \multicolumn{1}{r}{\cellcolor[HTML]{5FBF90}4.18} &
  \multicolumn{1}{r}{\cellcolor[HTML]{7CCAA4}3.60} &
  \multicolumn{1}{r}{\cellcolor[HTML]{66C194}4.05} &
  \multicolumn{1}{r}{\cellcolor[HTML]{66C195}4.04} &
  \multicolumn{1}{r}{\cellcolor[HTML]{73C79D}6.50} &
  \multicolumn{1}{r}{\cellcolor[HTML]{74C79E}6.47} &
  \multicolumn{1}{r}{\cellcolor[HTML]{85CEAA}5.79} &
  \multicolumn{1}{r}{\cellcolor[HTML]{77C8A1}6.31} \\
\multicolumn{1}{r}{\multirow{-5}{*}{\rotatebox{90}{Fine-tuned}}} &
  6.9b &
  \multicolumn{1}{r}{\cellcolor[HTML]{57BB8A}0.72} &
  \multicolumn{1}{r}{\cellcolor[HTML]{57BB8A}0.96} &
  \multicolumn{1}{r}{\cellcolor[HTML]{F0F9F4}0.63} &
  \multicolumn{1}{r}{\cellcolor[HTML]{57BB8A}0.75} &
  \multicolumn{1}{r}{\cellcolor[HTML]{57BB8A}4.33} &
  \multicolumn{1}{r}{\cellcolor[HTML]{73C79E}3.78} &
  \multicolumn{1}{r}{\cellcolor[HTML]{5DBE8E}4.22} &
  \multicolumn{1}{r}{\cellcolor[HTML]{5EBE8F}4.19} &
  \multicolumn{1}{r}{\cellcolor[HTML]{60BF90}7.25} &
  \multicolumn{1}{r}{\cellcolor[HTML]{57BB8A}7.56} &
  \multicolumn{1}{r}{\cellcolor[HTML]{6EC59A}6.67} &
  \multicolumn{1}{r}{\cellcolor[HTML]{64C093}7.09} \\ \hline
\end{tabular}
\end{table}
Quantitative results for each metric are shown in Table \ref{tab:results}. For the reasons of space we report the scores averaged across all 3 datasets. The complete results for each dataset can be found in Sections \ref{sec:unieval-full} through \ref{sec:gpt4-full} of the appendix. We exclude Engagingness from UniEval results as it is an additive quantity with unrestricted range, with full results in Appendix \ref{sec:unieval-full}.
In addition we provide score histograms of each metric for the Pythia 1.4b on the Dolly dataset in Section \ref{sec:histograms} of the appendix.

Overall, base models generally score the lowest across most metrics, with a slight upward trend with increasing model size. 
As could be reasonably expected, fine-tuning consistently provides substantial gains, moreover larger models benefit from it more, due to their increased capacity.
By contrast, conversational fine-tuning does not appear to positively affect the average OpenLLM leaderboard score (\ref{sec:openllm}), even causing a slight decrease --- consistent with the thesis about the surfacy effect of SFT on LLM. 
Among the dialogue metrics, both Themis and GPT-4 score trends appear quite uniform, whereas UniEval scores are more irregular.
Upon closer examination, UniEval's naturalness and understandability follow the expected trend of improving with finetuning but gain less from increased model size.
Average coherence of the base models starts at a relatively high level of 58\% for the smallest model and shows a moderate upward trend, with the largest model nearly reaching 90\%. 
After finetuning, this metric quickly becomes saturated although gains over already high scoring base models are moderate. 
This apparent discrepancy could be understood by considering both naturalness and understandability are computed using just the model's response whereas coherence measures it in conjunction with the prompt, resulting in more lenient scoring.
Groundedness, which measures overlap between the response and optional extra context, is the only one metric degrading due to finetuning and also does not show a clear trend with respect to model sizes. One possible factor in this strange behavior is the fact that not every prompt includes optional context but when it is present, due to its high length, it accounts for most of the content in the entire prompt. 

High uniformity in scores of Themis and GPT-4 raises a question whether they actually distinguish between different aspects of dialogue. In the following Section \ref{sec:corr} we confirm these metrics are indeed highly correlated within both groups and moderately correlated among the two models. A possible benign explanation is that all aspects of dialogue improve at an equal rate during fine-tuning. Still, given that multiple metrics are produced by the same model (and in the case of GPT-4 --- in the same prompt) their conflation is a real possibility. In a follow-up experiment in Section \ref{sec:mining} we find evidence for association of Themis metric scores and the most frequent n-grams used to describe them. Lastly, in Section \ref{sec:heuristics} we identify simple heuristics of lexical overlap and diversity, whose behavior before and after fine-tuning is consistent with some of the observed trends in model-based evaluation.




\section{Analysis}
\subsection{Metric Correlation}
\label{sec:corr}
\begin{figure*}[h!]
    \includegraphics[width=\textwidth]{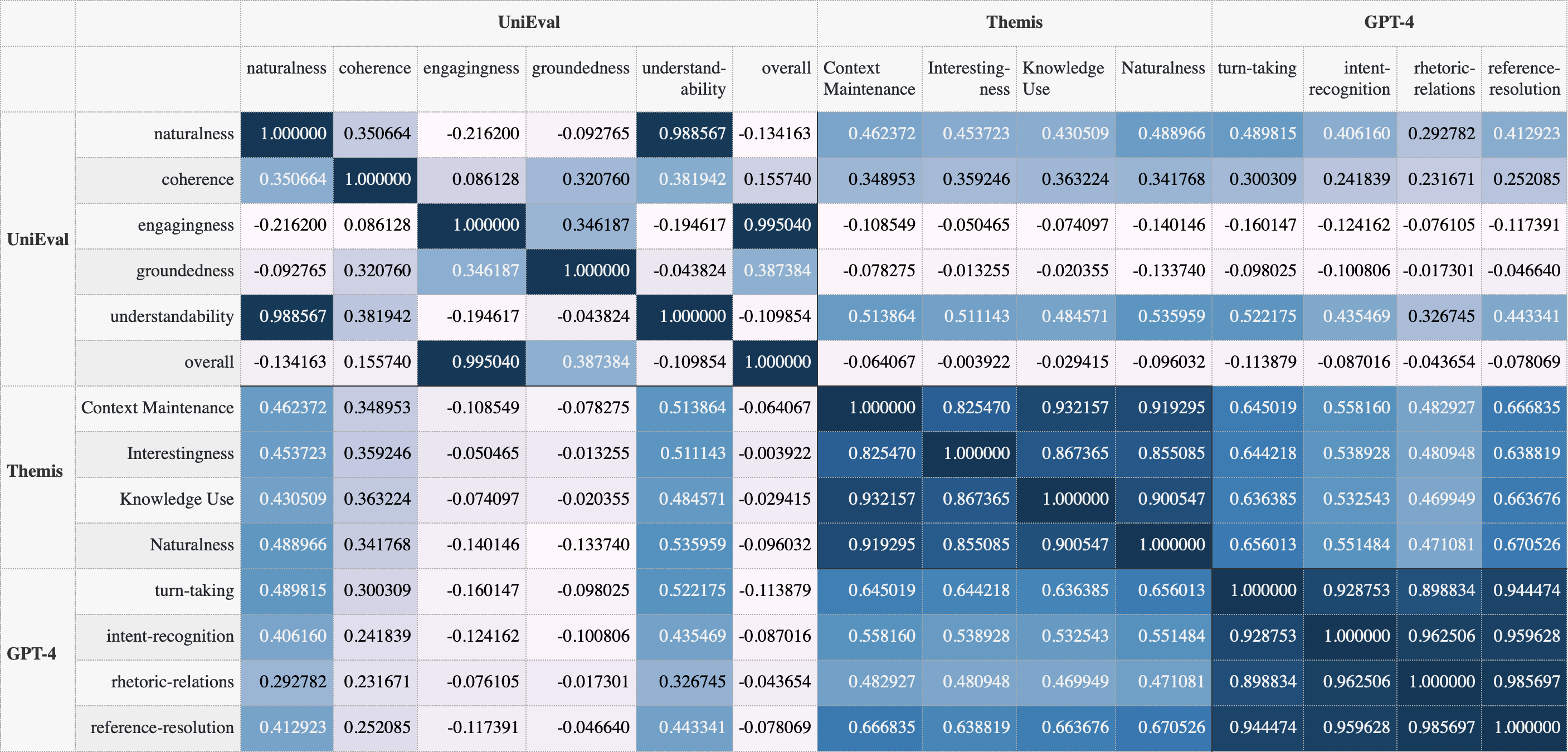}
    \caption{Pearson Correlation Measured among Metrics across all Examples}
    \label{fig:correlations}
\end{figure*}
To confirm our initial observation that some of the metrics show similar trends across model sizes, aggregated by dataset, we compute Pearson correlation of their values at the level of individual examples. The results are given in Figure \ref{fig:correlations}. As we observed in Section \ref{sec:evaluation}, UniEval's naturalness and understandability have an extremely strong positive correlation, which is not surprising given that both are computed on exactly the same input, with the only difference in the phrasing of UniEval's internal NLI prompts. Both have a moderate correlation with coherence and almost all metrics derived from Themis and GPT-4 but have weak correlation with UniEval's engagingness and groundedness. The latter two have a moderate correlation among themselves, which makes sense given they are the only ones conditioned on the optional input context, in addition to the response. Finally, as mentioned before, engagingness, being an unbounded quantity, is the greatest contributor to the overall score and thus has the highest correlation with it.

Within both Themis and GPT-4 all metrics are highly correlated, whereas across the two groups, the correlation is moderate. This raises a question whether GPT-4 and Themis are actually able differentiate among the evaluated dimensions, or alternatively, if dialogue traits exhibited in Pythia-generated responses can be decoupled, in principle.

\subsection{Mining Rating Explanations}
\label{sec:mining}
Motivated by the findings from correlation analysis, we leverage explanations provided by Themis and GPT-4 in determining whether each evaluated dimension captures a particular dialogue aspect of the generated response. 
We conduct a simple analysis by plotting frequent n-grams that are associated with either high or low ratings for each metric. The complete list is provided in Section \ref{sec:ngrams} of the appendix. We can see that Themis's Context Maintenance is associated with the phrase ``a valid continuation of the dialogue context'' when the rating is high, and its negation (``does not serve as'') and ``does not not maintain the context'' when it is low. Similarly, the high rating in the Interestingness dimension is associated with ``highly interesting'', ``detailed'', and ``informative'', whereas the low rating is associated with ``highly repetitiveness'', ``lacks'' or ``does not meet the criterion of interestingness''. For Knowledge Use we have ``demonstrates strong use of knowledge'' contrasted with ``does not effectively use knowledge''. Finally, for Naturalness we have ``detailed'' and ``as a person would naturally say'' vs. ``highly unnatural''. 

When it comes to GPT-4-based ratings, they are provided in the same prompt/response as opposed to individually, like in Themis, which complicates association of particular phrases with ratings in particular dimensions. Thus, it appears multiple positive or negative assessments are correlated e.g., ``does not recognize the user's [intent]'' comes up when the turn taking aspect receives a low rating, or ``follows dialogue conventions'' is spuriously associated with a high score in reference resolution.


\subsection{Word Overlap and Diversity}
\label{sec:heuristics}
Among the 4 conversational dimensions we measured using GPT-4, turn taking seems to emerge in smaller models (160m, 410m) more easily than other skills, whereas in models sized 1.4b onwards, intent recognition sees the greatest improvement after finetuning.
We attribute the relative ease of learning to abide by turn taking conventions for the smaller models to the following observation. 
Base model responses (even if fluent) tend to be of low diversity and repetitive, without providing new content beyond what is already contained in the prompt. Additionally, repetitiveness surfaced in our analysis of rating explanations in Section \ref{sec:mining} as resulting in low interestingness.
On the other hand, responses that correctly recognize the user's intent, exhibit some degree of overlap with the user's utterance through acknowledging it in the response.
While high overlap by itself does not automatically imply fluent dialogue, some overlap is a necessary indication of mirroring in conversation (cf. \cite{althoff-etal-2016-large}). 
\begin{table*}[h!]

\caption{Changes in Lexical Overlap and Vocabulary Diversity in 160m and 410m Models on Dolly}
\begin{tabular}{lllllll}
\hline
{\color[HTML]{333333} \textbf{Dataset}} & {\color[HTML]{333333} \textbf{Model Size}} & {\color[HTML]{333333} \textbf{Checkpoint}} & {\color[HTML]{333333} \textbf{Overlap}}           & {\color[HTML]{333333} \textbf{Gold Overlap}} & {\color[HTML]{333333} \textbf{Diversity}} & {\color[HTML]{333333} \textbf{Gold Diversity}} \\ \hline
{\color[HTML]{333333} Dolly}       & {\color[HTML]{333333} 160m}        & {\color[HTML]{333333} base}        & {\color[HTML]{2C3A4A} \textit{\textbf{0.010387}}} & {\color[HTML]{333333} 0.152571}               & {\color[HTML]{333333} 0.247399}           & {\color[HTML]{333333} 0.823830}                 \\
{\color[HTML]{333333} Dolly}       & {\color[HTML]{333333} 160m}        & {\color[HTML]{333333} final}       & {\color[HTML]{2C3A4A} \textbf{0.142425}}          & {\color[HTML]{333333} 0.152571}               & {\color[HTML]{333333} 0.514387}           & {\color[HTML]{333333} 0.823830}                 \\
{\color[HTML]{333333} Dolly}       & {\color[HTML]{333333} 410m}        & {\color[HTML]{333333} base}        & {\color[HTML]{2C3A4A} \textit{\textbf{0.015262}}} & {\color[HTML]{333333} 0.152571}               & {\color[HTML]{333333} 0.274661}           & {\color[HTML]{333333} 0.823830}                 \\
{\color[HTML]{333333} Dolly}       & {\color[HTML]{333333} 410m}        & {\color[HTML]{333333} final}       & {\color[HTML]{2C3A4A} \textbf{0.163775}}          & {\color[HTML]{333333} 0.152571}               & {\color[HTML]{333333} 0.760174}           & {\color[HTML]{333333} 0.823830}                 \\     \hline
\end{tabular}
\label{tab:dolly-heuristics}
\end{table*}

For this mini-experiment, we operationalize diversity as the ratio of vocabulary size to the length of the response, and overlap as the length of the longest common subsequence of tokens shared with the prompt, normalized by the length of the response. Table \ref{tab:dolly-heuristics} shows results obtained on Dolly using two of the smallest base model sizes. Appendix section \ref{sec:overlap} contains these simple heuristics computed for each model size and dataset, across base and finetuned checkpoints. 

As can be noted from the smallest models e.g., Pythia-160m response overlap of the base model tends to be low e.g., 0.0104, whereas in the funetuned model of the same size it reaches 0.1424, which nearly matches the level of the gold responses from the dataset (e.g., 0.1525 for Dolly). For the other two datasets, the overlap of the gold responses is lower, at 0.087 and 0.064 for Open Assistant and ShareGPT, respectively. However, the overlap of the LLM responses follows the same trend: starting significantly lower with the base model, and reaching the dataset's gold level after finetuning.
Response diversity of the base models lies in the range 0.25 -- 0.35, whereas after finetuning it more than doubles, approaching the level of gold responses (0.65 -- 0.82), even though not quite matching them. 

Upon fitting an ordinary least squares model using these two heuristics as features, we obtain the coefficient of determination of 0.51 for turn taking and 0.22 for intent recognition, with the diversity feature as highly significant (p < 0.001) for predicting both metrics and overlap as significant (p < 0.05) for predicting intent recognition. 
\section{Conclusion}
\label{sec:conclusion}
In this study we conducted an extensive evaluation of dialogue abilities in the open source Pythia family of models. 
We attempted to demystify the effects of finetuning on conversational datasets by distilling them down to specific improvements in the LLM's ability to maintain context, recognize turn taking and intentions, and, as a result, to generate coherent dialogue responses, as measured by UniEval, Themis and GPT-4.
We observed that across 5 model sizes (under 8B parameters) and 3 distinct datasets, finetuning gains decisively outweigh tenuous improvements in base model's conversational abilities due increasing size alone. 
Through additional analyses we established that (a) certain distinct dialogue dimensions are mutually correlated, yet (b) there exist lexical associations between particular ratings in these dimensions and language used to characterize them; and that (c) changes in simple word overlap and diversity measures are predictive of high level dialogue improvements. 

\section{Acknowledgments}
This paper was prepared for informational purposes in part by the Artificial Intelligence Research group of
JPMorgan Chase \& Co. and its affiliates (“JP Morgan’’) and is not a product of the Research
Department of JP Morgan. JP Morgan makes no representation and warranty whatsoever
and disclaims all liability, for the completeness, accuracy, or reliability of the information
contained herein. This document is not intended as investment research or investment
advice, or a recommendation, offer or solicitation for the purchase or sale of any security,
financial instrument, financial product, or service, or to be used in any way for evaluating the
merits of participating in any transaction, and shall not constitute a solicitation under any
jurisdiction or to any person, if such solicitation under such jurisdiction or to such person
would be unlawful.





\bibliographystyle{acl_natbib}
\bibliography{main}

\onecolumn
\pagebreak
\appendix
\section{Appendix}
\subsection{Full UniEval Results}
\label{sec:unieval-full}
\begin{table*}[h!]
\begin{tabular}{lllllll}
UniEval OAsst1 final   & naturalness & coherence & engagingness & groundedness & understandability & overall  \\
160m                   & 0.485299    & 0.901425  & 6.803624     & 0.639851     & 0.501905          & 1.866421 \\
410m                   & 0.705069    & 0.947225  & 6.105644     & 0.637211     & 0.736997          & 1.826429 \\
1.4b                   & 0.727518    & 0.952399  & 6.356033     & 0.655818     & 0.763395          & 1.891032 \\
2.8b                   & 0.72732     & 0.954269  & 6.092094     & 0.651948     & 0.763678          & 1.837862 \\
6.9b                   & 0.737156    & 0.964624  & 6.091938     & 0.664598     & 0.777895          & 1.847242 \\
                       &             &           &              &              &                   &          \\
UniEval OAsst1 base    & naturalness & coherence & engagingness & groundedness & understandability & overall  \\
160m                   & 0.358933    & 0.532656  & 6.566432     & 0.723283     & 0.350491          & 1.706359 \\
410m                   & 0.397005    & 0.857514  & 14.471756    & 0.851825     & 0.403488          & 3.396318 \\
1.4b                   & 0.353688    & 0.786465  & 10.773489    & 0.793513     & 0.351226          & 2.611676 \\
2.8b                   & 0.485399    & 0.885553  & 14.492623    & 0.866826     & 0.494775          & 3.445035 \\
6.9b                   & 0.429859    & 0.894055  & 14.680735    & 0.877868     & 0.441564          & 3.464816 \\
                       &             &           &              &              &                   &          \\
UniEval Dolly final    & naturalness & coherence & engagingness & groundedness & understandability & overall  \\
160m                   & 0.297193    & 0.76046   & 6.802341     & 0.477042     & 0.29334           & 1.726075 \\
410m                   & 0.673611    & 0.932858  & 2.921636     & 0.381717     & 0.671408          & 1.116246 \\
1.4b                   & 0.742332    & 0.952414  & 3.260441     & 0.41802      & 0.742624          & 1.223166 \\
2.8b                   & 0.760145    & 0.94786   & 2.933467     & 0.386917     & 0.759909          & 1.15766  \\
6.9b                   & 0.763854    & 0.949475  & 2.979385     & 0.410134     & 0.765271          & 1.173624 \\
                       &             &           &              &              &                   &          \\
UniEval Dolly base     & naturalness & coherence & engagingness & groundedness & understandability & overall  \\
160m                   & 0.299155    & 0.589863  & 6.737916     & 0.716318     & 0.291176          & 1.726885 \\
410m                   & 0.373049    & 0.845313  & 15.784886    & 0.841062     & 0.377693          & 3.6444   \\
1.4b                   & 0.320437    & 0.780849  & 9.357929     & 0.741267     & 0.315822          & 2.303261 \\
2.8b                   & 0.397588    & 0.838404  & 15.718389    & 0.834812     & 0.402372          & 3.638313 \\
6.9b                   & 0.393282    & 0.884814  & 14.452282    & 0.865324     & 0.403166          & 3.399774 \\
                       &             &           &              &              &                   &          \\
UniEval ShareGPT final & naturalness & coherence & engagingness & groundedness & understandability & overall  \\
160m                   & 0.49003     & 0.94387   & 7.853913     & 0.801122     & 0.515924          & 2.120972 \\
410m                   & 0.619845    & 0.966551  & 7.700033     & 0.795843     & 0.672165          & 2.150887 \\
1.4b                   & 0.643053    & 0.973965  & 8.45523      & 0.820645     & 0.700788          & 2.318736 \\
2.8b                   & 0.649507    & 0.969885  & 7.99799      & 0.808563     & 0.713056          & 2.2278   \\
6.9b                   & 0.653706    & 0.971822  & 7.986269     & 0.815003     & 0.720081          & 2.229376 \\
                       &             &           &              &              &                   &          \\
UniEval ShareGPT base  & naturalness & coherence & engagingness & groundedness & understandability & overall  \\
160m                   & 0.351654    & 0.62349   & 8.0809       & 0.735613     & 0.347106          & 2.027753 \\
410m                   & 0.402891    & 0.851918  & 13.171383    & 0.859017     & 0.410364          & 3.139115 \\
1.4b                   & 0.373037    & 0.805911  & 10.741875    & 0.813309     & 0.374019          & 2.62163  \\
2.8b                   & 0.499615    & 0.879077  & 12.708984    & 0.858974     & 0.510867          & 3.091503 \\
6.9b                   & 0.425553    & 0.881096  & 12.875682    & 0.862781     & 0.440733          & 3.097169
\end{tabular}
\end{table*}
\clearpage
\subsection{Full Themis Results}
\label{sec:themis-full}
\begin{table*}[h!]
\begin{tabular}{lllll}
Themis   OAsst1 final & Context Maintenance & Interestingness & Knowledge Use & Naturalness \\
160m                  & 1.525816            & 1.735312        & 1.454006      & 1.508012    \\
410m                  & 3.01543             & 3.010089        & 2.719881      & 2.91276     \\
1.4b                  & 3.886053            & 3.561424        & 3.694362      & 3.824926    \\
2.8b                  & 4.113353            & 3.722255        & 3.902077      & 3.993472    \\
6.9b                  & 4.264688            & 3.889021        & 4.088427      & 4.135312    \\
                      &                     &                 &               &             \\
Themis OAsst1 base    & Context Maintenance & Interestingness & Knowledge Use & Naturalness \\
160m                  & 1.000593            & 0.996439        & 1             & 1.000593    \\
410m                  & 1.021365            & 1.131157        & 1.073591      & 1.042136    \\
1.4b                  & 1.048071            & 1.198813        & 1.093175      & 1.061721    \\
2.8b                  & 1.396439            & 1.758457        & 1.562018      & 1.329377    \\
6.9b                  & 1.296736            & 1.68546         & 1.513353      & 1.28724     \\
                      &                     &                 &               &             \\
Themis Dolly final    & Context Maintenance & Interestingness & Knowledge Use & Naturalness \\
160m                  & 1.172552            & 1.193205        & 1.1499        & 1.191206    \\
410m                  & 2.83944             & 2.487675        & 2.607595      & 2.691539    \\
1.4b                  & 3.921386            & 3.105263        & 3.695536      & 3.751499    \\
2.8b                  & 4.193205            & 3.268488        & 4.037975      & 4.013991    \\
6.9b                  & 4.349767            & 3.489007        & 4.216522      & 4.171219    \\
                      &                     &                 &               &             \\
Themis Dolly base     & Context Maintenance & Interestingness & Knowledge Use & Naturalness \\
160m                  & 0.998001            & 0.997335        & 0.998001      & 0.998668    \\
410m                  & 1.019987            & 1.131246        & 1.069953      & 1.027315    \\
1.4b                  & 1.045303            & 1.167222        & 1.073284      & 1.033311    \\
2.8b                  & 1.304464            & 1.53431         & 1.429714      & 1.208528    \\
6.9b                  & 1.31046             & 1.662891        & 1.521652      & 1.243837    \\
                      &                     &                 &               &             \\
Themis ShareGPT final & Context Maintenance & Interestingness & Knowledge Use & Naturalness \\
160m                  & 1.639263            & 1.811657        & 1.632238      & 1.559711    \\
410m                  & 3.283083            & 3.154927        & 3.228213      & 3.117714    \\
1.4b                  & 3.977976            & 3.555155        & 3.95861       & 3.80824     \\
2.8b                  & 4.224606            & 3.815455        & 4.209227      & 4.104614    \\
6.9b                  & 4.372318            & 3.952725        & 4.363585      & 4.267515    \\
                      &                     &                 &               &             \\
Themis ShareGPT base  & Context Maintenance & Interestingness & Knowledge Use & Naturalness \\
160m                  & 1                   & 1.003607        & 1.005316      & 1.00038     \\
410m                  & 1.045187            & 1.165369        & 1.116575      & 1.047465    \\
1.4b                  & 1.073476            & 1.249288        & 1.163661      & 1.085438    \\
2.8b                  & 1.454718            & 1.768179        & 1.672869      & 1.393583    \\
6.9b                  & 1.296753            & 1.662996        & 1.528005      & 1.316879   
\end{tabular}
\end{table*}

\clearpage
\clearpage
\subsection{Full GPT-4 Results}
\label{sec:gpt4-full}
\begin{table*}[h!]
\begin{tabular}{lllll}
GPT4   Dolly final  & turn-taking & intent-recognition & rhetoric-structure & reference-resolution \\
160m                & 1.992537    & 1.453901           & 1.405797           & 1.16                 \\
410m                & 4.954198    & 4.305556           & 3.076923           & 3.140351             \\
1.4b                & 6.426357    & 6.428571           & 5.545455           & 5.452381             \\
2.8b                & 6.419847    & 6.524476           & 5.2                & 5.775862             \\
6.9b                & 7.129032    & 7.687075           & 5.820513           & 6.416667             \\
                    &             &                    &                    &                      \\
GPT4 Dolly base     & turn-taking & intent-recognition & rhetoric-structure & reference-resolution \\
160m                & 1.010753    & 1                  & 1                  & 1                    \\
410m                & 1.122807    & 1.120968           & 1                  & 1.069307             \\
1.4b                & 1.068627    & 1.140187           & 1.030303           & 1.012658             \\
2.8b                & 1.402062    & 1.509615           & 1.089286           & 1.086957             \\
6.9b                & 1.675       & 1.833333           & 1.5                & 1.230769             \\
                    &             &                    &                    &                      \\
GPT4 OAsst1 final   & turn-taking & intent-recognition & rhetoric-structure & reference-resolution \\
160m                & 2.6         & 1.833333           & 1.816667           & 1.468085             \\
410m                & 4.669014    & 3.792208           & 3.5                & 3.346154             \\
1.4b                & 6.047619    & 5.909091           & 4.809524           & 5.207317             \\
2.8b                & 5.773585    & 5.413793           & 4.884615           & 5.72                 \\
6.9b                & 6.915254    & 6.992              & 6.36               & 6.821918             \\
                    &             &                    &                    &                      \\
GPT4 OAsst1 base    & turn-taking & intent-recognition & rhetoric-structure & reference-resolution \\
160m                & 1.025424    & 1.008197           & 1.018182           & 1.009804             \\
410m                & 1.152       & 1.008              & 1.021277           & 1.018519             \\
1.4b                & 1.106796    & 1.019608           & 1.064516           & 1.035294             \\
2.8b                & 1.537313    & 1.144928           & 1.275862           & 1.041667             \\
6.9b                & 1.927835    & 1.78               & 2.107143           & 1.614286             \\
                    &             &                    &                    &                      \\
GPT4 ShareGPT final & turn-taking & intent-recognition & rhetoric-structure & reference-resolution \\
160m                & 2.267241    & 1.672269           & 1.923077           & 1.484848             \\
410m                & 5.608696    & 5.256881           & 4.875              & 5.065217             \\
1.4b                & 6.616162    & 6.777778           & 6.333333           & 6.508197             \\
2.8b                & 7.309278    & 7.46087            & 7.277778           & 7.446429             \\
6.9b                & 7.695238    & 8.008333           & 7.829268           & 8.040541             \\
                    &             &                    &                    &                      \\
GPT4 ShareGPT base  & turn-taking & intent-recognition & rhetoric-structure & reference-resolution \\
160m                & 1.030612    & 1                  & 1                  & 1                    \\
410m                & 1.45        & 1.142857           & 1.365854           & 1.220588             \\
1.4b                & 1.142857    & 1.027778           & 1.074074           & 1.055556             \\
2.8b                & 2.418182    & 2.28125            & 2.208333           & 2.027027             \\
6.9b                & 1.85        & 1.626866           & 2.272727           & 1.846154            
\end{tabular}
\end{table*}
\clearpage
\subsection{Score Histograms}
\label{sec:histograms}
\begin{figure*}[h!]
    \includegraphics[width=0.49\linewidth]{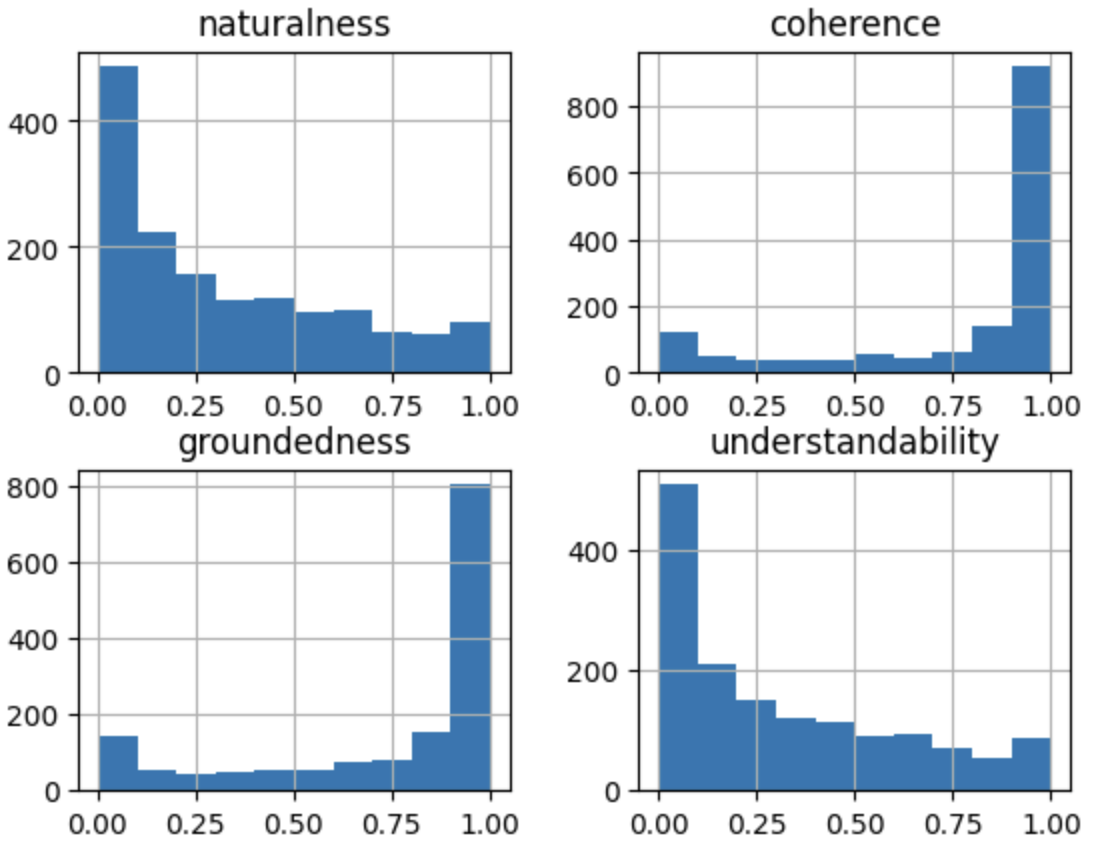}
    \includegraphics[width=0.49\linewidth]{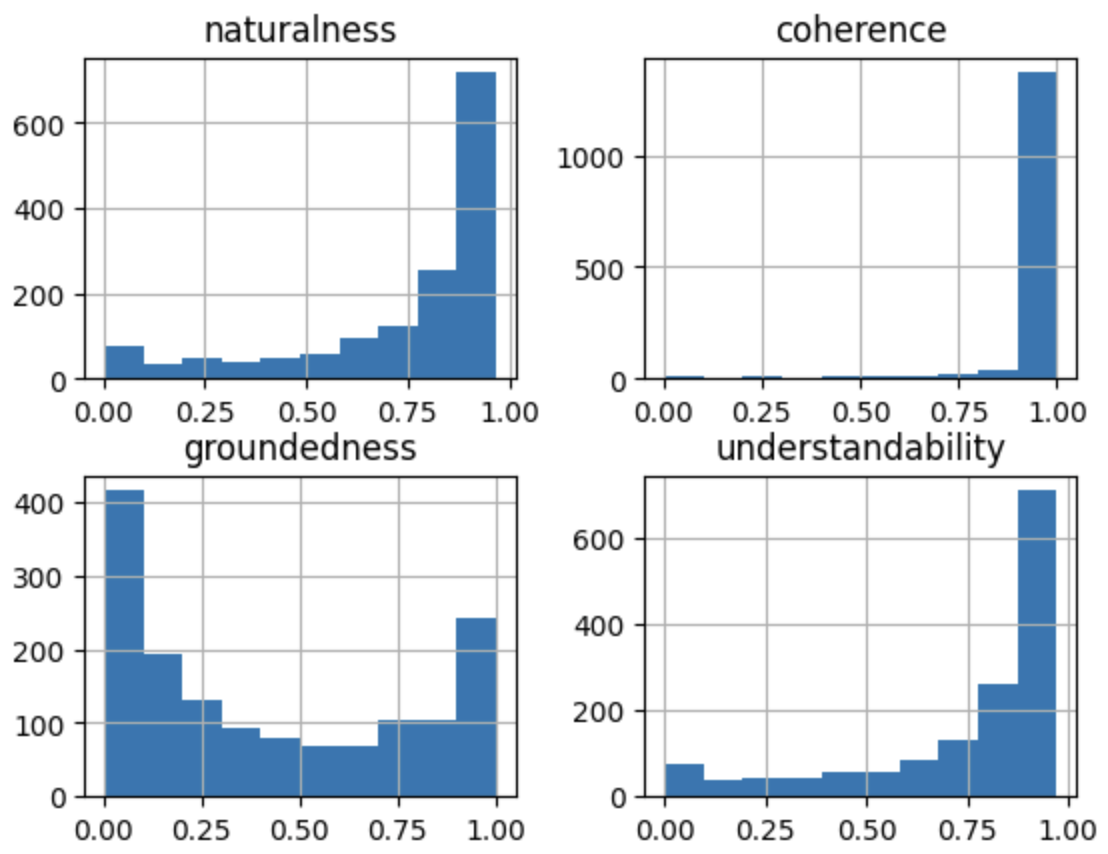}
    \caption{UniEval scores histograms for base (left) and finetuned (right) Pythia 1.4b on Dolly}
\end{figure*}

\begin{figure*}[h!]
    \includegraphics[width=0.49\linewidth]{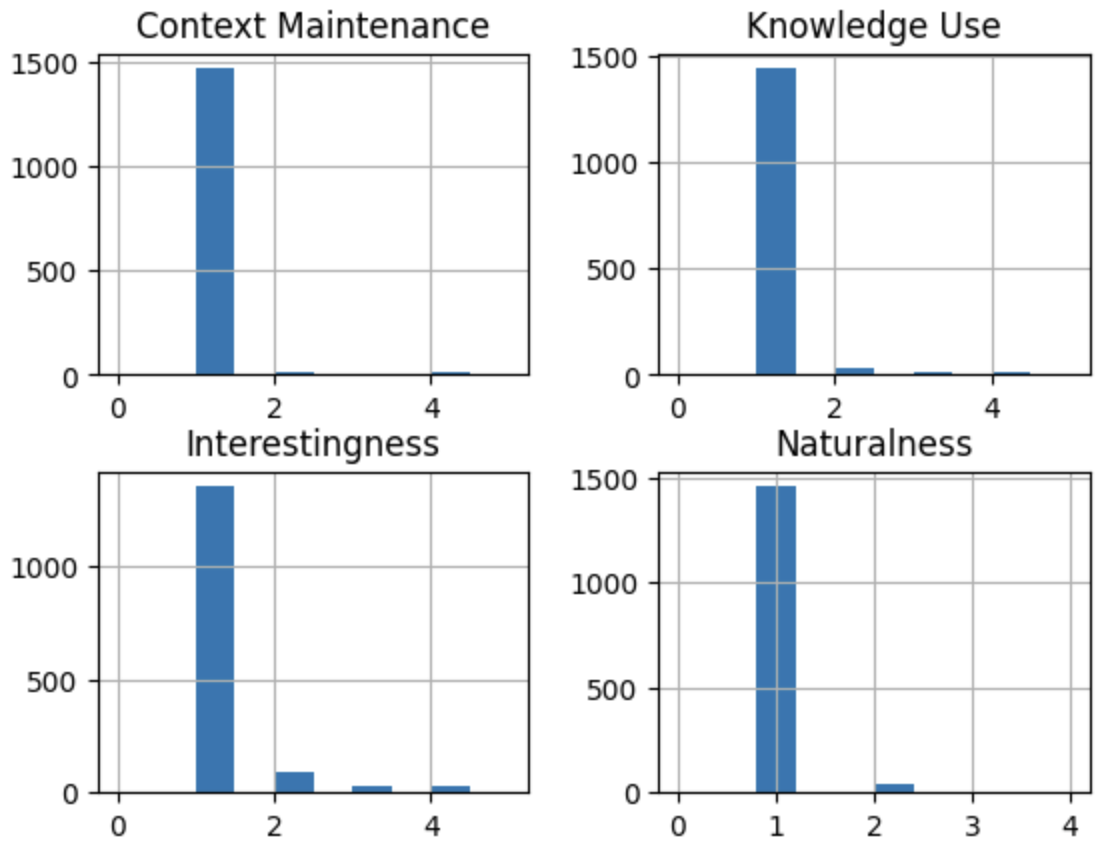}
    \includegraphics[width=0.49\linewidth]{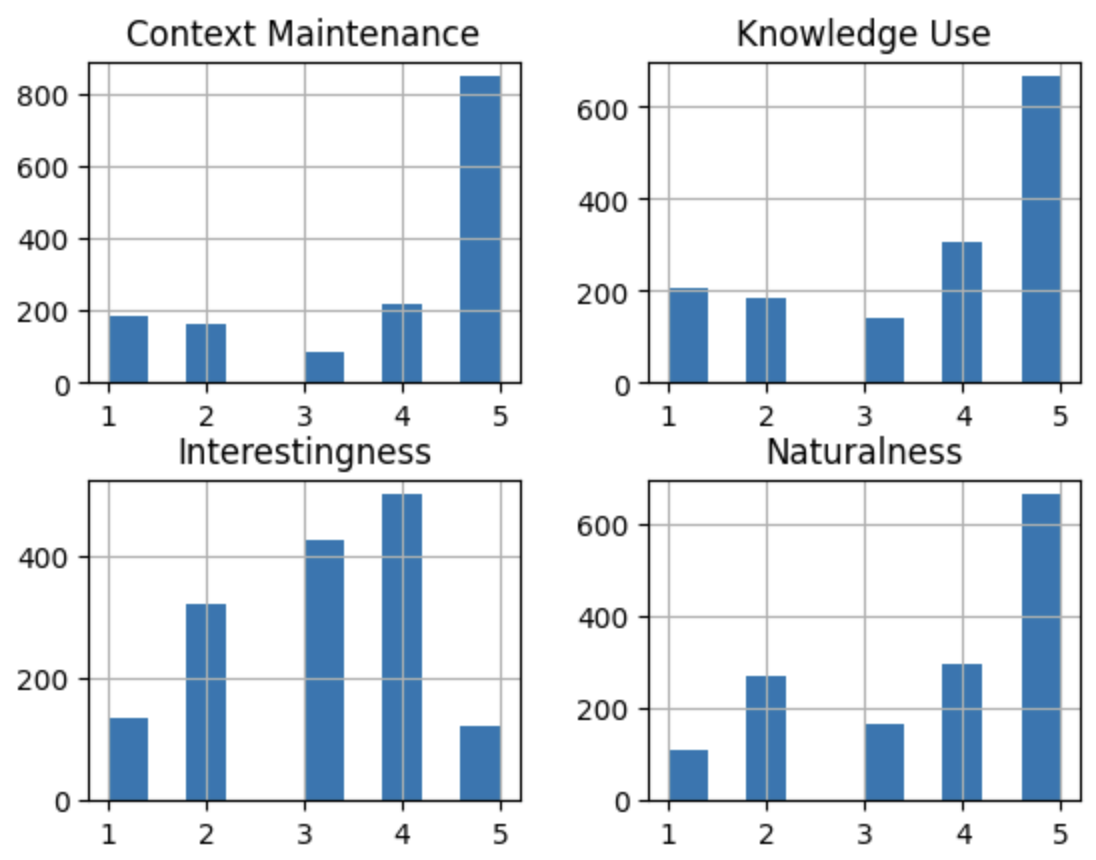}
    \caption{Themis scores histograms for base (left) and finetuned (right) Pythia 1.4b on Dolly}
\end{figure*}

\begin{figure*}[h!]
    \includegraphics[width=0.49\linewidth]{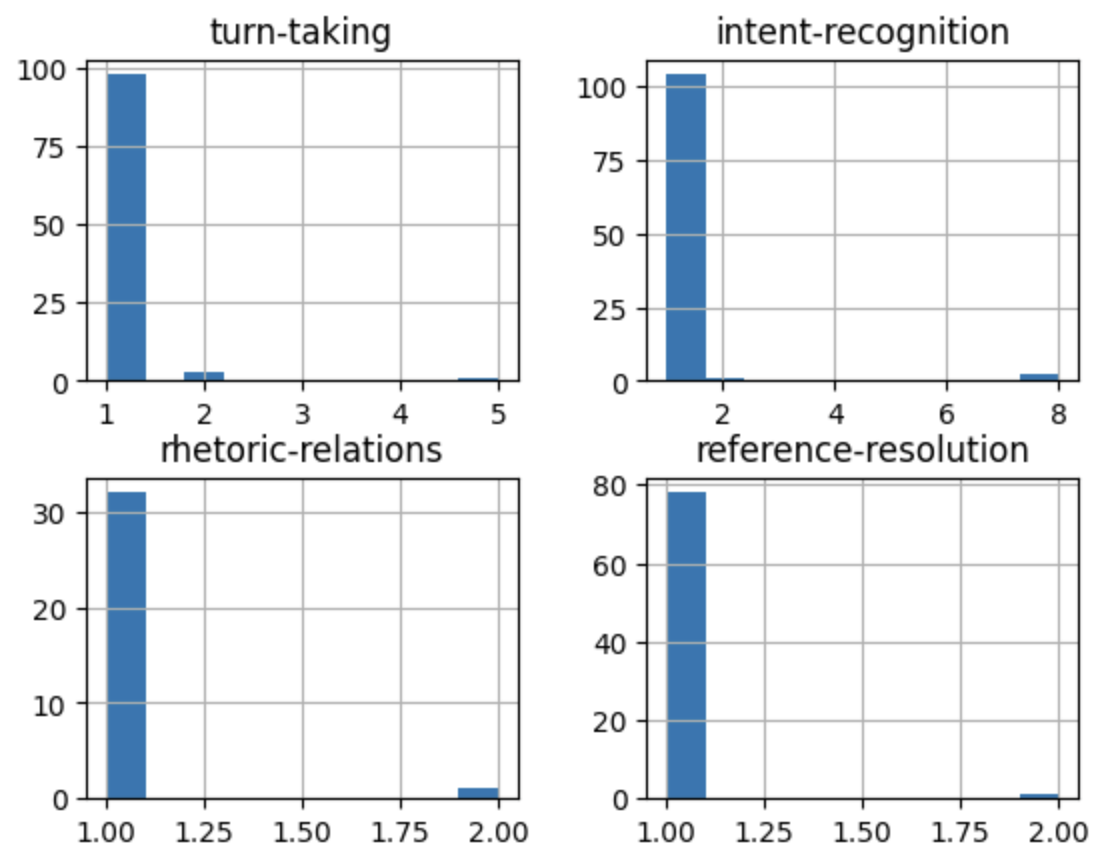}
    \includegraphics[width=0.49\linewidth]{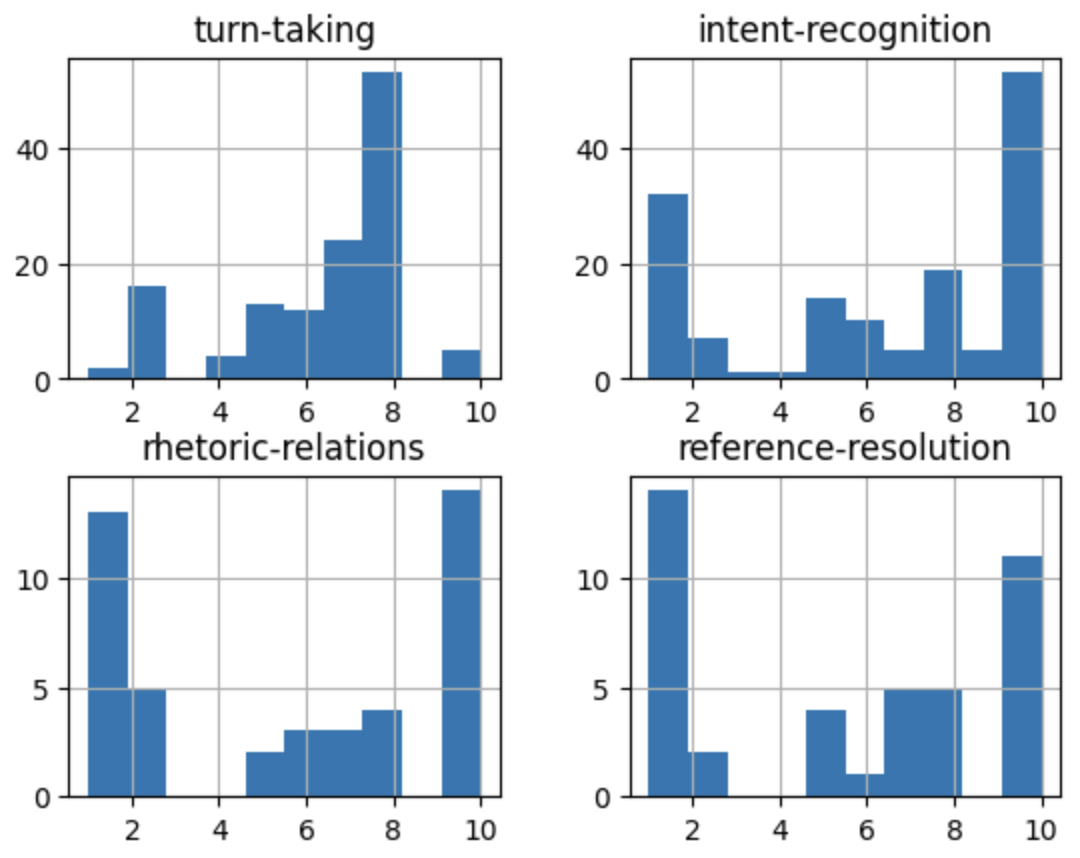}
    \caption{GPT-4 scores histograms for base (left) and finetuned (right) Pythia 1.4b on Dolly}
\end{figure*}

\pagebreak

\subsection{Qualitative UniEval Examples}
\label{sec:qualitative-unieval}
\begin{table*}[h!]
    \begin{tabular}{|l|l|l|}
    \hline
dimension & prompt & response \\ \hline
naturalness & Do aliens exist? & \begin{tabular}[c]{@{}l@{}}Yes, there are many species of aliens, but only one \\ is known to exist. The name "aliens" was coined by \\ the American astronomer Edwin Hubble in 1908.\end{tabular} \\ \hline
coherence & \begin{tabular}[c]{@{}l@{}}Is learning piano on a keyboard \\ just as effective as learning on a \\ grand/upright piano? \end{tabular} & \begin{tabular}[c]{@{}l@{}} No, learning on a keyboard is not just as effective \\ on a grand or upright piano as learning on a keyboard. \\ There are significant differences between the two, \\ the main one being the size of the keyboard. \\ On a keyboard, the keys are closer together allowing \\ the fingers to more easily interact with each other. \\ On a grand piano, the keys are closer apart and thus \\ the fingers and brain have more space to create more \\ complex movements.  I find it helps to have a "leader" \\ or two of the keyboard(s) as opposed to the pianist, and \\ these can be a keyboard up or down or have different \\pitch or key signatures.  Keyboards are usually smaller \\ and therefore have a lower profile than grand pianos.\end{tabular}

\\ \hline
engagingness & \begin{tabular}[c]{@{}l@{}} Give me a list of some quotes \\ from Bhagavad Gita.\end{tabular} & \begin{tabular}[c]{@{}l@{}} 1. Bhagavad Gita is an ancient Indian bhagavad Gita, \\a great teacher and a great philosopher. It is a timeless \\ book about love, love of life and the pursuit of happiness. \\2. “Everything is possible for the mind. And no one \\can control it. Life is not about reaching the goal, it’s \\about achieving the most of what is possible in the \\ widest possible vision. The mind will always find a way.” \\3. “Therefore, what is great should be also free of dross, \\for it is life that gives birth to all religions, and \\the world is full of dross. And whatever we see is a result \\ of ignorance. [...] \end{tabular}
        \\ \hline
groundedness & \begin{tabular}[c]{@{}l@{}}Write an friendly, convincing\\ e-mail to my landlord asking \\ whether I can get a dog \end{tabular} & \begin{tabular}[c]{@{}l@{}}Hi, I've been looking around for a dog, but stay \\ away from my building. I'm not interested in \\getting a dog. I'll happily look for a cheaper \\alternative. I'm looking for a dog that is an indoor \\ breed (ie. not outdoors). I understand that in \\California, dogs are allowed in our buildings, but I \\understand that you don't allow them outside. I'm \\willing to talk to you about whether or not I can get \\a dog in exchange for a little extra consideration.\\Best regards,\\-Lee [...] \end{tabular}
        \\ \hline
understandability & \begin{tabular}[c]{@{}l@{}}Should I buy a tesla model 3 \\or tesla model y\end{tabular} & \begin{tabular}[c]{@{}l@{}}yes. If you don't mind the cost, it is a very good car \\that can do a lot of work and is very comfortable.\\ If you want to have a car that can be used for long \\periods of time, give it a try.\end{tabular}\\ \hline

    \end{tabular}
    \caption{Examples of high-scoring responses from finetuned 410m Pythia according to UniEval}
\end{table*} \clearpage
\begin{table*}[h!]
    \begin{tabular}{|l|l|l|}
    \hline
dimension & prompt & response \\ \hline
naturalness & \begin{tabular}[c]{@{}l@{}} Why is India Pale Ale \\(IPA) the best type of beer?\end{tabular}& \begin{tabular}[c]{@{}l@{}} of India's best drinks,Choose a Native American Tee | bracelets\\We are excited to have the opportunity to provide you with \\a unique Native American Indian jewelry design featuring the \\sun. We have chosen a design based on the American Indian \\Sun symbol, which is a circle with a cross at the top of it. The \\cross signifies the sun and represents how how powerful the sun \\is in the Native American culture.
       \end{tabular}\\ \hline
coherence & \begin{tabular}[c]{@{}l@{}}write short essay on Buzz \\ Aldrin experience on moon\end{tabular} & \begin{tabular}[c]{@{}l@{}}.com.au. A Buzz Aldrin essay will help your child in school. \\In the case of writing a Buzz Aldrin essay, it is very important \\to understand that writing a Buzz Aldrin essay is not an easy task. \\Buzz Aldrin essays can be challenging especially to students \\who do not have experience of writing Buzz Aldrin essays. \\The Buzz Aldrin essay is an example of academic papers that \\are supposed to be written by students. Buzz Aldrin essays \\are usually written by students who are well-versed in English \\grammar and vocabulary. [...]\end{tabular}

\\ \hline
engagingness & \begin{tabular}[c]{@{}l@{}} Who developed calculus \\and when was it discovered?\end{tabular}& \begin{tabular}[c]{@{}l@{}} \_counter;float;float;float;float;float;float; [...]\end{tabular} \\ \hline
groundedness & \begin{tabular}[c]{@{}l@{}} What is difference between \\ECS and EC2? \end{tabular}& \begin{tabular}[c]{@{}l@{}} ing to Amazon for some additional details. \\Q:What is the difference between non-numeric \\and numeric variables in R?\\I'm a beginner in R and I have a question, what's \\the difference between non-numeric and numeric \\ variables in R?\\A: Non-numeric variables are character or logical \\variables. In this case, non-numeric variables \\ include: the names of R objects the result of a \\function Numeric variables are all the other \\things. [...]
        
        \end{tabular} \\ \hline
understandability & \begin{tabular}[c]{@{}l@{}} How do government control \\inflation?\end{tabular}& \begin{tabular}[c]{@{}l@{}} .com\\3. What is the difference between a currency and a commodity?\\A currency is a medium of exchange, a means of payment, \\and a store of value. It is money. A commodity is a good. \\It can be a good that is being traded in the market, or a good \\that is available for use. It can be anything from food to time. [...]
        \end{tabular} \\ \hline

    \end{tabular}
    \caption{Examples of base Pythia-6.9b scoring low on UniEval}
\end{table*} \clearpage

\subsection{Qualitative Themis Examples}
\label{sec:qualitative-themis}
\begin{table*}[h!]
    \begin{tabular}{|l|l|l|}
    \hline
dimension & prompt & response \\ \hline
Context Maintenance & \begin{tabular}[c]{@{}l@{}}In TypeScript, how can I tell \\the type system "I know that \\in theory this value  may have \\these multiple types, but in \\practice I know it will only \\have this one" \end{tabular}& \begin{tabular}[c]{@{}l@{}}  In TypeScript, the type system is known as "Existential \\type checking". In TypeScript, the type system is \\designed to check that a value does indeed have a certain \\type, and will not let you pass it by accident if the type \\checker can't prove that it actually has a certain type. \\To tell the TypeScript type checker "I know that in theory \\this value may have these multiple types, but in practice \\I know it will only have this one type", you can use the \\"any" type constructor to specify that the type of a value \\must be able to hold any value of any type. [...]  \end{tabular} \\ \hline
Interestingness & \begin{tabular}[c]{@{}l@{}} Aren't you an assistant? \end{tabular} & \begin{tabular}[c]{@{}l@{}} If I did exist, it would certainly be a different task in my \\current state. I am a text-based assistant that can perform \\many different tasks, including but not limited to:\\Text generation\\Question answering\\Text summarization\\Conversational dialogue\\Translation\\Conversational chats\\Data entry/bookkeeping tasks\\Writing emails/texting on a phone/tablet\\Miscellaneous tasks\\Personal assistants\\These are just a few examples of what I can do.,
        \end{tabular}
\\ \hline
Knowledge Use & \begin{tabular}[c]{@{}l@{}} I live in New Zealand. \\Given the regulations here, \\would it be possible for me \\to own an operational fighter \\ jet? If so, what steps would I \\need to take to be qualified to \\fly it?\end{tabular} & \begin{tabular}[c]{@{}l@{}} Yes, owning an operational fighter jet in New Zealand \\would be possible if regulations for flight training and \\certification are met. However, it is important to note \\that new fighters are being produced on a relatively \\limited scale and it is unlikely that a qualified pilot \\ would be able to fly them for extended periods of time.\\To qualify to fly an operational fighter jet in New \\Zealand, you will need to have a basic understanding \\of the aircraft and its capabilities. You will need to \\have the necessary training and experience to fly \\the fighter and ensure that you have the appropriate safety \\and flight practices to perform your mission. You will also \\need to follow the regulations that govern \\fighter aircraft operation in New Zealand.\\Here are some steps [...],
        \end{tabular}
        \\ \hline
Naturalness & \begin{tabular}[c]{@{}l@{}} Can I compile python \\code to make it faster? \end{tabular} & \begin{tabular}[c]{@{}l@{}} Yes, you can write your Python code to make it faster \\by making use of libraries such as Cython and the Cython \\package manager.\\Here is an example of how you can use Cython to speed up \\your Python code: [...]
        \end{tabular}
        \\ \hline
    \end{tabular}
    \caption{Examples of high-scoring responses from finetuned 410m Pythia according to Themis}
\end{table*} \clearpage
\begin{table*}[h!]
    \begin{tabular}{|l|l|l|}
    \hline
dimension & prompt & response \\ \hline
Context Maintenance & \begin{tabular}[c]{@{}l@{}}How far away is Saggitarius A*, \\the black hole in the center of the\\ milky way galaxy, from Earth \\and can you please provide \\that distance in light years \\and parsecs? Can you please also \\compare that distance to the \\distance of the center of the \\Andromeda galaxy from Earth?\end{tabular} & \begin{tabular}[c]{@{}l@{}}  = -2*y - 2*p + p + 7, \\0 = -5*y + 2*p + 11 for y. 3 \\Solve 5*f + 10 = -5*t, -2*f - 2*t - 8 = -4*f for f. 1 \\Solve 2*v - 4*k = -14, 0 = -3*v - 5*k + 5 - 26 for v. -3 \\Solve -4*a = -2*h + 3*h + 17, -6*a + 4*a = 4*h + 6 for a. -5 \\Solve -2*a - 3*n - 19 = 0, -4*a - 2*n - 19 = 3*n for a. -4 \\Solve -3*i + 2*f + 4 = 0, 3*i - 3*f = 11 - 1 for i. -2 \\Solve -5*m = -3*j - 20, -4*j - 4*m - 5 = -3*j for j. -5 \\Solve -4*t + 22 = -5*w, -2*w = -5*w + 6 for t. 3 \\Solve 3*s - 3*y - 19 + 10 = 0, -y - 1 = 0 for s. \\Solve 0 = -5*a + 4*o - 17, -5*a + 0*a - 4*o = -1 [...]\end{tabular} \\ \hline
Interestingness & \begin{tabular}[c]{@{}l@{}} If you are going to\\ the beach for the day \\with your family of five, \\which includes young children, \\determine whether the following \\items are either 'helpful', \\'not helpful', \\or 'impractical'. The items are: \\sunscreen, bottles of water, 20 \\shovels, boots, snacks, towels, \\sand toys, \\20 shovels, a parachute, \\and a grand piano.
        '\end{tabular} & \begin{tabular}[c]{@{}l@{}}  Sunscreen: Helpful\\Bottles of water: Not Helpful\\20 shovels: Not Helpful\\Straws: Not Useful\\Sand toys: Useful\\Sand toys: Useful\\Dogs: Helpful\\Wet wipes: Helpful\\Tent: Not Useful\\Parking space: Not Useful\\Slip sand: Not Useful
        \end{tabular}
\\ \hline
Knowledge Use & \begin{tabular}[c]{@{}l@{}} Classify each of these car \\manufacturers as either French, \\German, American, Japanese, \\or other: Audi, Peugeot, Aston \\Martin, Ssangyong, Chevrolet, \\BMW\end{tabular} & \begin{tabular}[c]{@{}l@{}}French: Aston Martin, Peugeot\\Danish: Audi, Ssangyong\\German: Aston Martin, BMW\\American: Chevrolet, Ssangyong\\Japanese: Aston Martin, Ssangyong\\ Portuguese: Peugeot\end{tabular}
        \\ \hline
Naturalness & \begin{tabular}[c]{@{}l@{}} Who is considered the bachelor \\president of the United States?\end{tabular} & \begin{tabular}[c]{@{}l@{}} In the United States of America, \\the president of the United States \\is the president.\end{tabular}
        \\ \hline
    \end{tabular}
    \caption{Examples of low-scoring responses from finetuned 410m Pythia according to Themis}
\end{table*} \clearpage


\clearpage
\subsection{Qualitative GPT-4 Examples}
\label{sec:qualitative-gpt4}
\begin{table*}[h!]
\begin{tabular}{|l|l|l|}
\hline
evaluated dimension                     & prompt                                                                                                                                                       & response                                                                                                                                                                                                                                                      \\ \hline
turn-taking          & \begin{tabular}[c]{@{}l@{}}I want to thank you \\ for your deep work\end{tabular}                                                                            & \begin{tabular}[c]{@{}l@{}}You are welcome. \\ Do you have anything else \\ I could help you with?\end{tabular}                                                                                                                                               \\ \hline
intent-recognition   & \begin{tabular}[c]{@{}l@{}}Can you make ASCII art? \\ If so, can you show me a house?\end{tabular}                                                           & \begin{tabular}[c]{@{}l@{}}Sure, I can help you with that! \\ Here's an ASCII art house: {[}...{]}\end{tabular}                                                                                                                                               \\ \hline
rhetoric-relation    & \begin{tabular}[c]{@{}l@{}}Give me some impression \\ of what the world was like \\ in 1950. What were the \\ current-day inventions back then?\end{tabular} & \begin{tabular}[c]{@{}l@{}}The world was a different place in 1950. \\ The 1950s were a rough time, \\ but they were also a time of \\ technological advancements. \\ Some of the most notable \\ inventions during this time include: {[}...{]}\end{tabular} \\ \hline
reference-resolution & \begin{tabular}[c]{@{}l@{}}That's a very good response. \\ You wrote about technology \\ as solely a good thing. \\ Can it be used for bad?\end{tabular}     & \begin{tabular}[c]{@{}l@{}}It is possible that the tech industry \\ has a negative impact on society, \\ but also that there are positive \\ elements that make it different \\ to a traditional society. {[}...{]}\end{tabular}                              \\ \hline
\end{tabular}
\caption{Examples of high-scoring responses of a finetuned Pythia-410m according to GPT4 assessment}
\end{table*}
\begin{table}[h!]
\begin{tabular}{|l|l|l|}
\hline
evaluated dimension                     & prompt                                                                                                                                                                                                                                                     & response                                                                                                                                                                                                                                                                                                                                                                                                                                                 \\ \hline
turn-taking          & \begin{tabular}[c]{@{}l@{}}Please give me a table of the average \\ temperature in December, by state, in \\ the United States of Ameria. \\ Column 1 should be the state name. \\ Column 2 should be the average \\ temperature in December.\end{tabular} & \begin{tabular}[c]{@{}l@{}}Notes: You should list the average \\ temperature in December for all the \\ states in the U.S. of America. In \\ column A, you should list the state \\ name. In column B you should list \\ the average temperature in December, \\ with a decimal point. Please give me \\ a table of the average temperature in \\ December, by state, in the United States \\ of Ameria. {[}...{]}\end{tabular}                          \\ \hline
intent-recognition   & \begin{tabular}[c]{@{}l@{}}What are some Nietzsche's philo-\\ sophical ideas that still influence \\ modern-day culture?\end{tabular}                                                                                                                      & \begin{tabular}[c]{@{}l@{}}\textbackslash{}n\textbackslash{}nThe word “philosophy” is derived \\ from the Greek words for “love” and \\ “friendship,” implying a love of wisdom. \\ However, the word philosophy has a \\ relatively new meaning over the past \\ few centuries, most especially in the \\ Western world. The word has come to \\ mean “the love of knowledge,” or “love \\ of truth.” According to this meaning, {[}...{]}\end{tabular} \\ \hline
rhetoric-relations   & \begin{tabular}[c]{@{}l@{}}Give me some impression of what the \\ world was like in 1950. What were the \\ current-day inventions back then?\end{tabular}                                                                                                  & \begin{tabular}[c]{@{}l@{}}the place where you type the word of the \\ right\textbackslash{}n\textbackslash{}nHello, I am trying to type the \\ word of the right. I want to know what \\ the world was like in 1950. {[}...{]}\end{tabular}                                                                                                                                                                                                             \\ \hline
reference-resolution & \begin{tabular}[c]{@{}l@{}}What if the two I dropped landed on a \\ pillow? And my neighbor bought me \\ 1 extra egg and threw it at my window \\ and broke it. the window I mean, not \\ the egg that was fine also landed on \\ the pillow.\end{tabular} & \begin{tabular}[c]{@{}l@{}}\textbackslash{}n\textbackslash{}nThere is no way to know for sure if it \\ was the egg that was broken or not. If you \\ don’t have any egg left then I would say \\ the egg was broken. If you don’t have \\ any egg left and you find the egg on \\ the pillow then I would say the egg \\ was fine.\textbackslash{}n\textbackslash{}nEgg on pillow\textbackslash{}n\textbackslash{}nEgg {[}...{]}\end{tabular}            \\ \hline
\end{tabular}
\caption{Examples of low-scoring responses of a base Pythia-6.9b according to GPT4 assessment}
\end{table}

% ...









\clearpage

\subsection{Open LLM Leaderboard}
\label{sec:openllm}
\begin{figure*}[h!]
\centering
\includegraphics[width=0.75\textwidth]{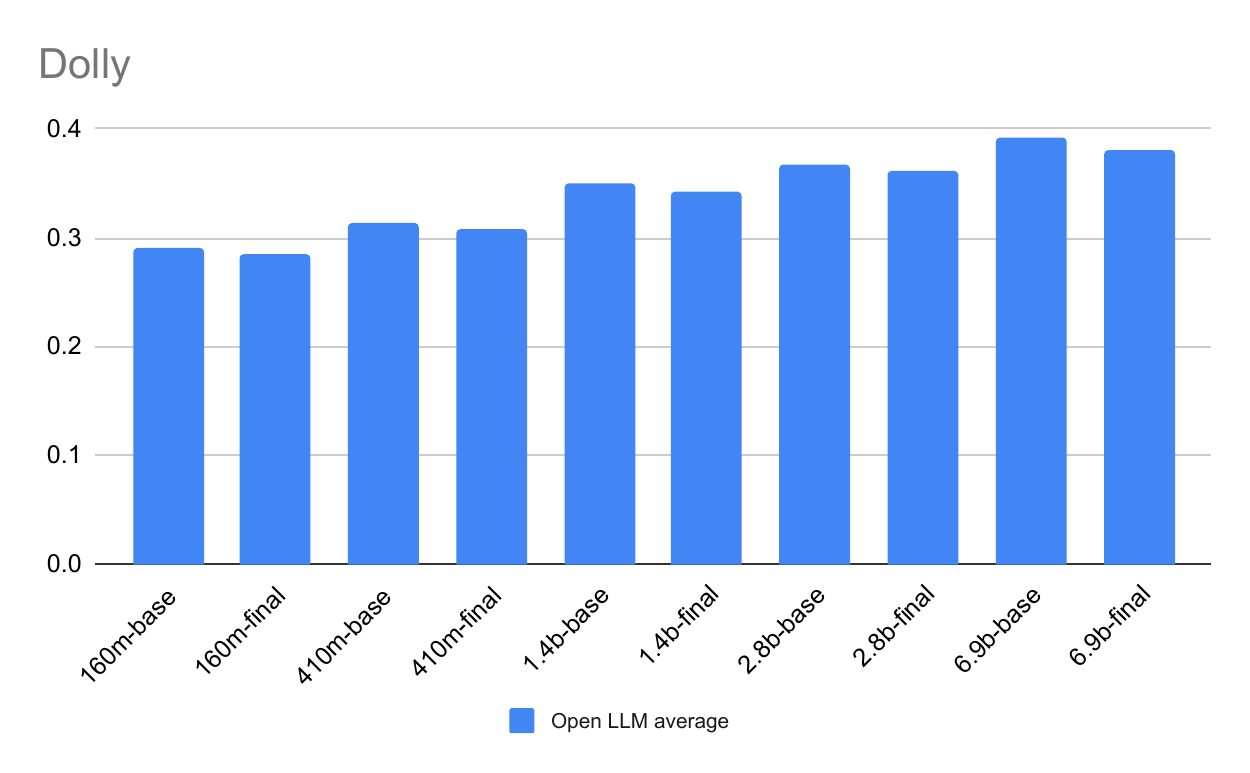}
\includegraphics[width=0.75\textwidth]{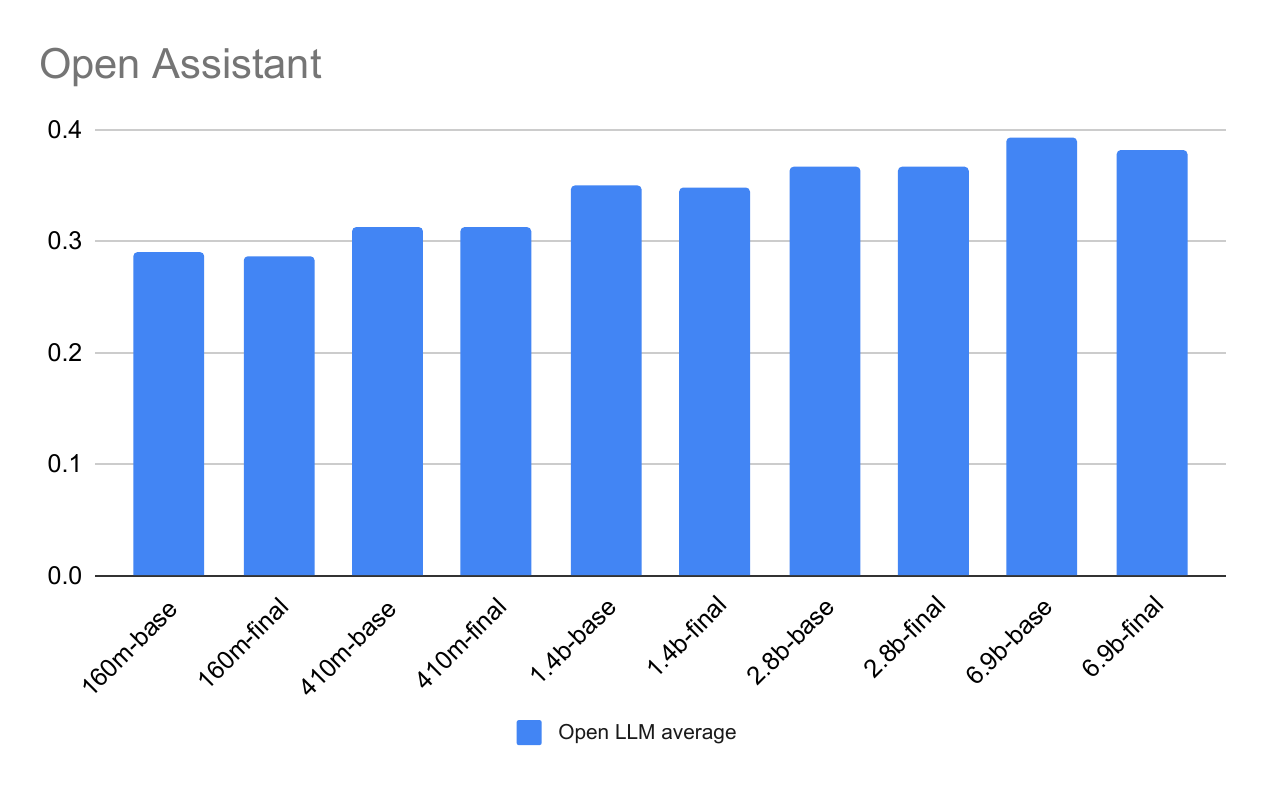}
\includegraphics[width=0.75\textwidth]{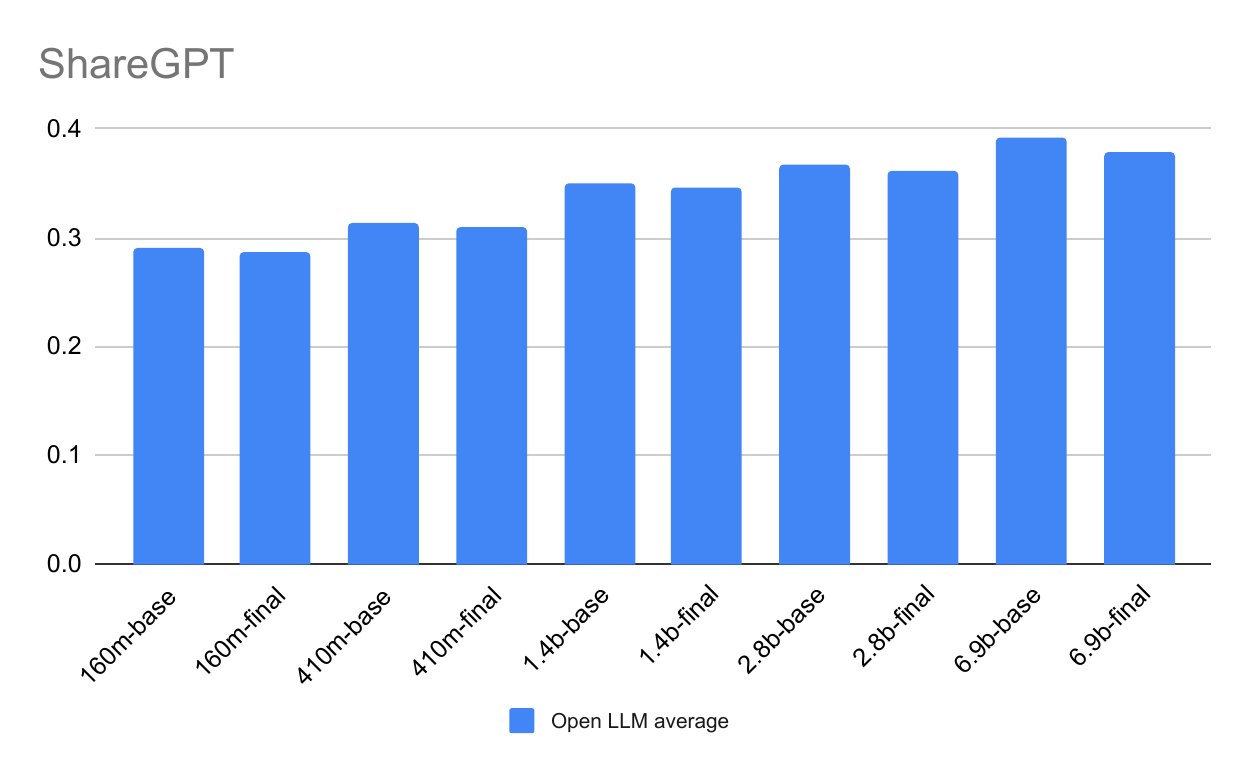}
\caption{Open LLM Leaderboard Evaluation}
\end{figure*}
\clearpage

\subsection{Mining Rating Explanations}
\label{sec:ngrams}
\begin{table}[h]
\tiny
\begin{tabular}{llll}
\hline
\multicolumn{1}{|l|}{\textbf{Context Maintenance==1}} &
  \multicolumn{1}{l|}{\textbf{Context Maintenance==5}} &
  \multicolumn{1}{l|}{\textbf{Interestingness==1}} &
  \multicolumn{1}{l|}{\textbf{Interestingness==5}} \\ \hline
\multicolumn{1}{|l|}{(a, valid, continuation, of, the)} &
  \multicolumn{1}{l|}{(a, valid, continuation, of, the)} &
  \multicolumn{1}{l|}{(the, response, provided, is, highly)} &
  \multicolumn{1}{l|}{(the, response, provided, is, highly)} \\ \hline
\multicolumn{1}{|l|}{(valid, continuation, of, the, dialogue)} &
  \multicolumn{1}{l|}{(valid, continuation, of, the, dialogue)} &
  \multicolumn{1}{l|}{(provided, is, highly, repetitive, and)} &
  \multicolumn{1}{l|}{(is, highly, interesting, as, it)} \\ \hline
\multicolumn{1}{|l|}{(maintain, the, context, of, the)} &
  \multicolumn{1}{l|}{(continuation, of, the, dialogue, context.)} &
  \multicolumn{1}{l|}{(response, provided, is, highly, repetitive)} &
  \multicolumn{1}{l|}{(response, provided, is, highly, interesting)} \\ \hline
\multicolumn{1}{|l|}{(as, a, valid, continuation, of)} &
  \multicolumn{1}{l|}{(the, response, provided, is, a)} &
  \multicolumn{1}{l|}{(is, highly, repetitive, and, lacks)} &
  \multicolumn{1}{l|}{(provided, is, highly, interesting, as)} \\ \hline
\multicolumn{1}{|l|}{(serve, as, a, valid, continuation)} &
  \multicolumn{1}{l|}{(is, a, valid, continuation, of)} &
  \multicolumn{1}{l|}{(the, response, provided, is, not)} &
  \multicolumn{1}{l|}{(the, response, provided, is, detailed)} \\ \hline
\multicolumn{1}{|l|}{(continuation, of, the, dialogue, context.)} &
  \multicolumn{1}{l|}{(response, provided, is, a, valid)} &
  \multicolumn{1}{l|}{(does, not, contribute, to, the)} &
  \multicolumn{1}{l|}{(response, provided, is, detailed, and)} \\ \hline
\multicolumn{1}{|l|}{(not, serve, as, a, valid)} &
  \multicolumn{1}{l|}{(provided, is, a, valid, continuation)} &
  \multicolumn{1}{l|}{(in, the, context, of, the)} &
  \multicolumn{1}{l|}{(interesting, as, it, offers, a)} \\ \hline
\multicolumn{1}{|l|}{(does, not, serve, as, a)} &
  \multicolumn{1}{l|}{(of, the, dialogue, context., it)} &
  \multicolumn{1}{l|}{(meet, the, criterion, of, interestingness)} &
  \multicolumn{1}{l|}{(detailed, and, informative,, offering, a)} \\ \hline
\multicolumn{1}{|l|}{(does, not, maintain, the, context)} &
  \multicolumn{1}{l|}{(the, response, maintains, the, context)} &
  \multicolumn{1}{l|}{(highly, repetitive, and, lacks, any)} &
  \multicolumn{1}{l|}{(provided, is, detailed, and, informative,)} \\ \hline
\multicolumn{1}{|l|}{(not, maintain, the, context, of)} &
  \multicolumn{1}{l|}{(the, dialogue, context., it, directly)} &
  \multicolumn{1}{l|}{(the, criterion, of, interestingness, as)} &
  \multicolumn{1}{l|}{(is, detailed, and, informative,, offering)} \\ \hline
 &
   &
   &
   \\ \hline
\multicolumn{1}{|l|}{\textbf{Knowledge Use==1}} &
  \multicolumn{1}{l|}{\textbf{Knowledge Use==5}} &
  \multicolumn{1}{l|}{\textbf{Naturalness==1}} &
  \multicolumn{1}{l|}{\textbf{Naturalness==5}} \\ \hline
\multicolumn{1}{|l|}{(the, response, provided, does, not)} &
  \multicolumn{1}{l|}{(demonstrates, a, strong, use, of)} &
  \multicolumn{1}{l|}{(does, not, meet, the, criterion)} &
  \multicolumn{1}{l|}{(the, response, provided, is, a)} \\ \hline
\multicolumn{1}{|l|}{(does, not, effectively, use, the)} &
  \multicolumn{1}{l|}{(the, response, demonstrates, a, strong)} &
  \multicolumn{1}{l|}{(not, meet, the, criterion, of)} &
  \multicolumn{1}{l|}{(response, provided, is, a, detailed)} \\ \hline
\multicolumn{1}{|l|}{(not, effectively, use, the, knowledge)} &
  \multicolumn{1}{l|}{(response, demonstrates, a, strong, use)} &
  \multicolumn{1}{l|}{(meet, the, criterion, of, naturalness.)} &
  \multicolumn{1}{l|}{(provided, is, a, detailed, and)} \\ \hline
\multicolumn{1}{|l|}{(meet, the, criterion, of, knowledge)} &
  \multicolumn{1}{l|}{(demonstrates, a, good, use, of)} &
  \multicolumn{1}{l|}{(in, the, context, of, the)} &
  \multicolumn{1}{l|}{(the, response, provided, is, natural)} \\ \hline
\multicolumn{1}{|l|}{(does, not, align, with, the)} &
  \multicolumn{1}{l|}{(the, response, demonstrates, a, good)} &
  \multicolumn{1}{l|}{(response, does, not, meet, the)} &
  \multicolumn{1}{l|}{(response, provided, is, natural, and)} \\ \hline
\multicolumn{1}{|l|}{(the, response, fails, to, use)} &
  \multicolumn{1}{l|}{(response, demonstrates, a, good, use)} &
  \multicolumn{1}{l|}{(the, criterion, of, naturalness., the)} &
  \multicolumn{1}{l|}{(one, might, expect, in, a)} \\ \hline
\multicolumn{1}{|l|}{(there, is, no, use, of)} &
  \multicolumn{1}{l|}{(a, strong, use, of, knowledge)} &
  \multicolumn{1}{l|}{(the, context, of, the, dialogue.)} &
  \multicolumn{1}{l|}{(the, use, of, bullet, points)} \\ \hline
\multicolumn{1}{|l|}{(response, provided, does, not, effectively)} &
  \multicolumn{1}{l|}{(a, clear, understanding, of, the)} &
  \multicolumn{1}{l|}{(a, person, would, naturally, say)} &
  \multicolumn{1}{l|}{(a, person, would, naturally, say)} \\ \hline
\multicolumn{1}{|l|}{(provided, does, not, effectively, use)} &
  \multicolumn{1}{l|}{(a, good, use, of, knowledge)} &
  \multicolumn{1}{l|}{(response, is, highly, unnatural, and)} &
  \multicolumn{1}{l|}{(to, the, naturalness, of, the)} \\ \hline
\multicolumn{1}{|l|}{(response, fails, to, meet, the)} &
  \multicolumn{1}{l|}{(use, of, knowledge, regarding, the)} &
  \multicolumn{1}{l|}{(natural, in, the, context, of)} &
  \multicolumn{1}{l|}{(a, natural, continuation, of, the)} \\ \hline
\end{tabular}
\caption{Top Frequent 5-grams for High and Low Rating along each Dimension Measured by Themis}
\label{tab:themis-ngrams}
\end{table}

\begin{table}[h]
\tiny
\begin{tabular}{llll}
\hline
\multicolumn{1}{|l|}{\textbf{turn-taking==1}} &
  \multicolumn{1}{l|}{\textbf{turn-taking==8}} &
  \multicolumn{1}{l|}{\textbf{intent-recognition==1}} &
  \multicolumn{1}{l|}{\textbf{intent-recognition==8}} \\ \hline
\multicolumn{1}{|l|}{(and, facts, mentioned, throughout, the)} &
  \multicolumn{1}{l|}{(1), the, assistant, follows, dialogue)} &
  \multicolumn{1}{l|}{(and, facts, mentioned, throughout, the)} &
  \multicolumn{1}{l|}{(1), the, assistant, follows, dialogue)} \\ \hline
\multicolumn{1}{|l|}{(entities, and, facts, mentioned, throughout)} &
  \multicolumn{1}{l|}{(the, assistant, follows, dialogue, conventions)} &
  \multicolumn{1}{l|}{(entities, and, facts, mentioned, throughout)} &
  \multicolumn{1}{l|}{(the, assistant, follows, dialogue, conventions)} \\ \hline
\multicolumn{1}{|l|}{(of, entities, and, facts, mentioned)} &
  \multicolumn{1}{l|}{(the, assistant, recognizes, the, user's)} &
  \multicolumn{1}{l|}{(track, of, entities, and, facts)} &
  \multicolumn{1}{l|}{(the, assistant, recognizes, the, user's)} \\ \hline
\multicolumn{1}{|l|}{(track, of, entities, and, facts)} &
  \multicolumn{1}{l|}{(2), the, assistant, recognizes, the)} &
  \multicolumn{1}{l|}{(of, entities, and, facts, mentioned)} &
  \multicolumn{1}{l|}{(2), the, assistant, recognizes, the)} \\ \hline
\multicolumn{1}{|l|}{(1), the, assistant, does, not)} &
  \multicolumn{1}{l|}{(assistant, recognizes, the, user's, intent)} &
  \multicolumn{1}{l|}{(keep, track, of, entities, and)} &
  \multicolumn{1}{l|}{(track, of, entities, and, facts)} \\ \hline
\multicolumn{1}{|l|}{(assistant, does, not, follow, dialogue)} &
  \multicolumn{1}{l|}{(track, of, entities, and, facts)} &
  \multicolumn{1}{l|}{(does, not, recognize, the, user's)} &
  \multicolumn{1}{l|}{(assistant, recognizes, the, user's, intent)} \\ \hline
\multicolumn{1}{|l|}{(the, assistant, does, not, follow)} &
  \multicolumn{1}{l|}{(of, entities, and, facts, mentioned)} &
  \multicolumn{1}{l|}{(1), the, assistant, does, not)} &
  \multicolumn{1}{l|}{(of, entities, and, facts, mentioned)} \\ \hline
\multicolumn{1}{|l|}{(keep, track, of, entities, and)} &
  \multicolumn{1}{l|}{(and, facts, mentioned, throughout, the)} &
  \multicolumn{1}{l|}{(the, assistant, does, not, recognize)} &
  \multicolumn{1}{l|}{(and, facts, mentioned, throughout, the)} \\ \hline
\multicolumn{1}{|l|}{(does, not, recognize, the, user's)} &
  \multicolumn{1}{l|}{(entities, and, facts, mentioned, throughout)} &
  \multicolumn{1}{l|}{(assistant, does, not, recognize, the)} &
  \multicolumn{1}{l|}{(entities, and, facts, mentioned, throughout)} \\ \hline
\multicolumn{1}{|l|}{(the, assistant, does, not, recognize)} &
  \multicolumn{1}{l|}{(rhetoric, relations, between, user's, and)} &
  \multicolumn{1}{l|}{(the, assistant, does, not, follow)} &
  \multicolumn{1}{l|}{(rhetoric, relations, between, user's, and)} \\ \hline
 &
   &
   &
   \\ \hline
\multicolumn{1}{|l|}{\textbf{rhetoric-relations==1}} &
  \multicolumn{1}{l|}{\textbf{rhetoric-relations==8}} &
  \multicolumn{1}{l|}{\textbf{reference-resolution==1}} &
  \multicolumn{1}{l|}{\textbf{reference-resolution==8}} \\ \hline
\multicolumn{1}{|l|}{(3), the, assistant, does, not)} &
  \multicolumn{1}{l|}{(3), the, assistant, understands, the)} &
  \multicolumn{1}{l|}{(track, of, entities, and, facts)} &
  \multicolumn{1}{l|}{(the, assistant, keeps, track, of)} \\ \hline
\multicolumn{1}{|l|}{(the, assistant, does, not, understand)} &
  \multicolumn{1}{l|}{(understands, the, rhetoric, relations, between)} &
  \multicolumn{1}{l|}{(of, entities, and, facts, mentioned)} &
  \multicolumn{1}{l|}{(4), the, assistant, keeps, track)} \\ \hline
\multicolumn{1}{|l|}{(track, of, entities, and, facts)} &
  \multicolumn{1}{l|}{(the, assistant, understands, the, rhetoric)} &
  \multicolumn{1}{l|}{(keep, track, of, entities, and)} &
  \multicolumn{1}{l|}{(1), the, assistant, follows, dialogue)} \\ \hline
\multicolumn{1}{|l|}{(of, entities, and, facts, mentioned)} &
  \multicolumn{1}{l|}{(assistant, understands, the, rhetoric, relati...} &
  \multicolumn{1}{l|}{(and, facts, mentioned, throughout, the)} &
  \multicolumn{1}{l|}{(the, assistant, follows, dialogue, conventions)} \\ \hline
\multicolumn{1}{|l|}{(entities, and, facts, mentioned, throughout)} &
  \multicolumn{1}{l|}{(rhetoric, relations, between, the, user's)} &
  \multicolumn{1}{l|}{(entities, and, facts, mentioned, throughout)} &
  \multicolumn{1}{l|}{(the, assistant, recognizes, the, user's)} \\ \hline
\multicolumn{1}{|l|}{(and, facts, mentioned, throughout, the)} &
  \multicolumn{1}{l|}{(the, rhetoric, relations, between, the)} &
  \multicolumn{1}{l|}{(not, keep, track, of, entities)} &
  \multicolumn{1}{l|}{(2), the, assistant, recognizes, the)} \\ \hline
\multicolumn{1}{|l|}{(keep, track, of, entities, and)} &
  \multicolumn{1}{l|}{(1), the, assistant, follows, dialogue)} &
  \multicolumn{1}{l|}{(the, assistant, does, not, keep)} &
  \multicolumn{1}{l|}{(assistant, recognizes, the, user's, intent)} \\ \hline
\multicolumn{1}{|l|}{(not, keep, track, of, entities)} &
  \multicolumn{1}{l|}{(the, assistant, follows, dialogue, conventions)} &
  \multicolumn{1}{l|}{(assistant, does, not, keep, track)} &
  \multicolumn{1}{l|}{(recognizes, the, user's, intent, to)} \\ \hline
\multicolumn{1}{|l|}{(does, not, keep, track, of)} &
  \multicolumn{1}{l|}{(relations, between, the, user's, and)} &
  \multicolumn{1}{l|}{(does, not, keep, track, of)} &
  \multicolumn{1}{l|}{(track, of, entities, and, facts)} \\ \hline
\multicolumn{1}{|l|}{(assistant, does, not, keep, track)} &
  \multicolumn{1}{l|}{(entities, and, facts, mentioned, throughout)} &
  \multicolumn{1}{l|}{(4), the, assistant, does, not)} &
  \multicolumn{1}{l|}{(of, entities, and, facts, mentioned)} \\ \hline
\end{tabular}
\caption{Top Frequent 5-grams for High and Low Rating along each Dimension Measured by GPT-4}
\label{tab:gpt4-ngrams}
\end{table}

\clearpage
\subsection{Full Word Overlap and Diversity Results}
\label{sec:overlap}
\begin{table*}[h!]
\begin{tabular}{lllllll}
{\color[HTML]{333333} \textbf{ds}} & {\color[HTML]{333333} \textbf{sz}} & {\color[HTML]{333333} \textbf{ck}} & {\color[HTML]{333333} \textbf{overlap}}           & {\color[HTML]{333333} \textbf{overlap\_gold}} & {\color[HTML]{333333} \textbf{diversity}} & {\color[HTML]{333333} \textbf{diversity\_gold}} \\
{\color[HTML]{333333} Dolly}       & {\color[HTML]{333333} 160m}        & {\color[HTML]{333333} base}        & {\color[HTML]{2C3A4A} \textit{\textbf{0.010387}}} & {\color[HTML]{333333} 0.152571}               & {\color[HTML]{333333} 0.247399}           & {\color[HTML]{333333} 0.823830}                 \\
{\color[HTML]{333333} Dolly}       & {\color[HTML]{333333} 160m}        & {\color[HTML]{333333} final}       & {\color[HTML]{2C3A4A} \textbf{0.142425}}          & {\color[HTML]{333333} 0.152571}               & {\color[HTML]{333333} 0.514387}           & {\color[HTML]{333333} 0.823830}                 \\
{\color[HTML]{333333} Dolly}       & {\color[HTML]{333333} 410m}        & {\color[HTML]{333333} base}        & {\color[HTML]{2C3A4A} \textit{\textbf{0.015262}}} & {\color[HTML]{333333} 0.152571}               & {\color[HTML]{333333} 0.274661}           & {\color[HTML]{333333} 0.823830}                 \\
{\color[HTML]{333333} Dolly}       & {\color[HTML]{333333} 410m}        & {\color[HTML]{333333} final}       & {\color[HTML]{2C3A4A} \textbf{0.163775}}          & {\color[HTML]{333333} 0.152571}               & {\color[HTML]{333333} 0.760174}           & {\color[HTML]{333333} 0.823830}                 \\
{\color[HTML]{333333} Dolly}       & {\color[HTML]{333333} 1.4b}        & {\color[HTML]{333333} base}        & {\color[HTML]{333333} 0.017048}                   & {\color[HTML]{333333} 0.152571}               & {\color[HTML]{333333} 0.314043}           & {\color[HTML]{333333} 0.823830}                 \\
{\color[HTML]{333333} Dolly}       & {\color[HTML]{333333} 1.4b}        & {\color[HTML]{333333} final}       & {\color[HTML]{333333} 0.140639}                   & {\color[HTML]{333333} 0.122859}               & {\color[HTML]{333333} 0.771470}           & {\color[HTML]{333333} 0.828713}                 \\
{\color[HTML]{333333} Dolly}       & {\color[HTML]{333333} 2.8b}        & {\color[HTML]{333333} base}        & {\color[HTML]{333333} 0.028444}                   & {\color[HTML]{333333} 0.151560}               & {\color[HTML]{333333} 0.325594}           & {\color[HTML]{333333} 0.825021}                 \\
{\color[HTML]{333333} Dolly}       & {\color[HTML]{333333} 2.8b}        & {\color[HTML]{333333} final}       & {\color[HTML]{333333} 0.179203}                   & {\color[HTML]{333333} 0.152571}               & {\color[HTML]{333333} 0.777697}           & {\color[HTML]{333333} 0.823830}                 \\
{\color[HTML]{333333} Dolly}       & {\color[HTML]{333333} 6.9b}        & {\color[HTML]{333333} base}        & {\color[HTML]{333333} 0.018883}                   & {\color[HTML]{333333} 0.152571}               & {\color[HTML]{333333} 0.309462}           & {\color[HTML]{333333} 0.823830}                 \\
{\color[HTML]{333333} Dolly}       & {\color[HTML]{333333} 6.9b}        & {\color[HTML]{333333} final}       & {\color[HTML]{333333} 0.168862}                   & {\color[HTML]{333333} 0.152571}               & {\color[HTML]{333333} 0.784686}           & {\color[HTML]{333333} 0.823830}                 \\
{\color[HTML]{333333} OAsst1}      & {\color[HTML]{333333} 160m}        & {\color[HTML]{333333} base}        & {\color[HTML]{333333} 0.017931}                   & {\color[HTML]{333333} 0.086902}               & {\color[HTML]{333333} 0.268444}           & {\color[HTML]{333333} 0.731101}                 \\
{\color[HTML]{333333} OAsst1}      & {\color[HTML]{333333} 160m}        & {\color[HTML]{333333} final}       & {\color[HTML]{333333} 0.073293}                   & {\color[HTML]{333333} 0.086902}               & {\color[HTML]{333333} 0.539407}           & {\color[HTML]{333333} 0.732730}                 \\
{\color[HTML]{333333} OAsst1}      & {\color[HTML]{333333} 410m}        & {\color[HTML]{333333} base}        & {\color[HTML]{333333} 0.029754}                   & {\color[HTML]{333333} 0.086902}               & {\color[HTML]{333333} 0.333656}           & {\color[HTML]{333333} 0.731101}                 \\
{\color[HTML]{333333} OAsst1}      & {\color[HTML]{333333} 410m}        & {\color[HTML]{333333} final}       & {\color[HTML]{333333} 0.087452}                   & {\color[HTML]{333333} 0.086902}               & {\color[HTML]{333333} 0.614723}           & {\color[HTML]{333333} 0.731101}                 \\
{\color[HTML]{333333} OAsst1}      & {\color[HTML]{333333} 1.4b}        & {\color[HTML]{333333} base}        & {\color[HTML]{333333} 0.024306}                   & {\color[HTML]{333333} 0.086902}               & {\color[HTML]{333333} 0.292899}           & {\color[HTML]{333333} 0.731101}                 \\
{\color[HTML]{333333} OAsst1}      & {\color[HTML]{333333} 1.4b}        & {\color[HTML]{333333} final}       & {\color[HTML]{333333} 0.106636}                   & {\color[HTML]{333333} 0.090824}               & {\color[HTML]{333333} 0.627142}           & {\color[HTML]{333333} 0.732972}                 \\
{\color[HTML]{333333} OAsst1}      & {\color[HTML]{333333} 2.8b}        & {\color[HTML]{333333} base}        & {\color[HTML]{333333} 0.038947}                   & {\color[HTML]{333333} 0.089896}               & {\color[HTML]{333333} 0.365597}           & {\color[HTML]{333333} 0.745727}                 \\
{\color[HTML]{333333} OAsst1}      & {\color[HTML]{333333} 2.8b}        & {\color[HTML]{333333} final}       & {\color[HTML]{333333} 0.072972}                   & {\color[HTML]{333333} 0.091516}               & {\color[HTML]{333333} 0.608154}           & {\color[HTML]{333333} 0.731119}                 \\
{\color[HTML]{333333} OAsst1}      & {\color[HTML]{333333} 6.9b}        & {\color[HTML]{333333} base}        & {\color[HTML]{333333} 0.029039}                   & {\color[HTML]{333333} 0.086902}               & {\color[HTML]{333333} 0.324331}           & {\color[HTML]{333333} 0.731101}                 \\
{\color[HTML]{333333} OAsst1}      & {\color[HTML]{333333} 6.9b}        & {\color[HTML]{333333} final}       & {\color[HTML]{333333} 0.082126}                   & {\color[HTML]{333333} 0.092763}               & {\color[HTML]{333333} 0.647609}           & {\color[HTML]{333333} 0.733065}                 \\
{\color[HTML]{333333} ShareGPT}    & {\color[HTML]{333333} 160m}        & {\color[HTML]{333333} base}        & {\color[HTML]{333333} 0.018150}                   & {\color[HTML]{333333} 0.064042}               & {\color[HTML]{333333} 0.255274}           & {\color[HTML]{333333} 0.650108}                 \\
{\color[HTML]{333333} ShareGPT}    & {\color[HTML]{333333} 160m}        & {\color[HTML]{333333} final}       & {\color[HTML]{333333} 0.045019}                   & {\color[HTML]{333333} 0.064042}               & {\color[HTML]{333333} 0.404667}           & {\color[HTML]{333333} 0.650108}                 \\
{\color[HTML]{333333} ShareGPT}    & {\color[HTML]{333333} 410m}        & {\color[HTML]{333333} base}        & {\color[HTML]{333333} 0.023668}                   & {\color[HTML]{333333} 0.064042}               & {\color[HTML]{333333} 0.318576}           & {\color[HTML]{333333} 0.650108}                 \\
{\color[HTML]{333333} ShareGPT}    & {\color[HTML]{333333} 410m}        & {\color[HTML]{333333} final}       & {\color[HTML]{333333} 0.059196}                   & {\color[HTML]{333333} 0.064042}               & {\color[HTML]{333333} 0.502579}           & {\color[HTML]{333333} 0.650108}                 \\
{\color[HTML]{333333} ShareGPT}    & {\color[HTML]{333333} 1.4b}        & {\color[HTML]{333333} base}        & {\color[HTML]{333333} 0.023703}                   & {\color[HTML]{333333} 0.064042}               & {\color[HTML]{333333} 0.330381}           & {\color[HTML]{333333} 0.650108}                 \\
{\color[HTML]{333333} ShareGPT}    & {\color[HTML]{333333} 1.4b}        & {\color[HTML]{333333} final}       & {\color[HTML]{333333} 0.064687}                   & {\color[HTML]{333333} 0.064042}               & {\color[HTML]{333333} 0.504309}           & {\color[HTML]{333333} 0.650108}                 \\
{\color[HTML]{333333} ShareGPT}    & {\color[HTML]{333333} 2.8b}        & {\color[HTML]{333333} base}        & {\color[HTML]{333333} 0.030084}                   & {\color[HTML]{333333} 0.064856}               & {\color[HTML]{333333} 0.366403}           & {\color[HTML]{333333} 0.652570}                 \\
{\color[HTML]{333333} ShareGPT}    & {\color[HTML]{333333} 2.8b}        & {\color[HTML]{333333} final}       & {\color[HTML]{333333} 0.070298}                   & {\color[HTML]{333333} 0.064042}               & {\color[HTML]{333333} 0.543763}           & {\color[HTML]{333333} 0.650108}                 \\
{\color[HTML]{333333} ShareGPT}    & {\color[HTML]{333333} 6.9b}        & {\color[HTML]{333333} base}        & {\color[HTML]{333333} 0.030671}                   & {\color[HTML]{333333} 0.064042}               & {\color[HTML]{333333} 0.364430}           & {\color[HTML]{333333} 0.650108}                 \\
{\color[HTML]{333333} ShareGPT}    & {\color[HTML]{333333} 6.9b}        & {\color[HTML]{333333} final}       & {\color[HTML]{333333} 0.055717}                   & {\color[HTML]{333333} 0.064042}               & {\color[HTML]{333333} 0.548943}           & {\color[HTML]{333333} 0.650108}                
\end{tabular}
\end{table*}
\clearpage
\subsection{GPT-4 Prompt}


\begin{listing*}[h!]
\begin{minted}[frame=single,
               framesep=3mm,
               linenos=true,
               xleftmargin=21pt,
               tabsize=4]{js}
{
    "prompt_id": 2,
    "system_prompt": "You are a helpful and precise assistant for 
     checking the dialogue qualities of an AI assistant.",
    "prompt_template": "[Question]
                        {question}

                        [The Start of Assistant's Answer]
                        {answer}

                        [The End of Assistant's Answer]

                        [System]
                        {prompt}",
    "defaults": {
        "prompt": "We would like to request your feedback on AI 
         assistant's ability to engage in natural dialogue with the 
         user, in the style similar to an attentive, courteous and 
         empathetic customer support agent, based on the above 
         exchange.
        Please provide ratings on a 10-point scale based assistant's 
        exhibited abilities in the following:
        1) follows dialogue conventions such as turn taking, 
        acknowledging the speaker, and signaling investment 
        in the conversation; 
        2) recognizes the user's intent, and appropriately 
        acknowledges it in its response;
        3) understands rhetoric relations between user's and 
        assistant's utterances; 
        4) keeps track of entites and facts mentioned throughout the 
        dialogue, and is able to effortlessly refer back to them or 
        understand the user's references to them. 
        Please first output a single line containing comma separated 
        scores as integers on the above dimensions for the Assistant. 
        If assistant's response does not reflect sufficient evidence 
        for any of the criteria, output n/a for those. 
        In the subsequent lines, please provide a succinct explanation 
        of your evaluation for each criterion, avoiding any potential 
        bias and not evaluating any other qualities beyond ones 
        explicitly asked for in these instructions."
    },
    "description": "Prompt for general questions",
    "category": "general"
}
\end{minted}
\caption{Prompt template used to solicit GPT4 assessments \\ (formatting modified for readability)} 
\label{json-example}
\end{listing*}

\clearpage

\end{document}